\documentclass{article}

\PassOptionsToPackage{table}{xcolor}
\PassOptionsToPackage{numbers,sort&compress}{natbib}
\usepackage[preprint]{neurips_2026}

\usepackage[utf8]{inputenc} 
\usepackage[T1]{fontenc}    
\usepackage[colorlinks=true, citecolor=blue, linkcolor=black, urlcolor=black]{hyperref}
\usepackage{titletoc}       
\usepackage{url}            
\usepackage{booktabs}       
\usepackage{amsfonts}       
\usepackage{nicefrac}       
\usepackage{microtype}      
\usepackage{xcolor}         

\usepackage{graphicx}
\usepackage{tcolorbox}
\usepackage{tikz}
\usetikzlibrary{positioning,arrows.meta,calc}
\usepackage{pgfplots}
\usepackage{bm}
\usepgfplotslibrary{groupplots}
\pgfplotsset{compat=1.18}
\usepackage[accsupp]{axessibility} 
\usepackage{enumitem}
\usepackage{soul}
\usepackage{orcidlink}
\usepackage{caption}
\usepackage{subcaption}
\usepackage{xspace}
\usepackage{listings}
\usepackage{placeins}
\usepackage{alltt}
\usepackage[capitalize]{cleveref}

\makeatletter
\DeclareRobustCommand\onedot{\futurelet\@let@token\@onedot}
\def\@onedot{\ifx\@let@token.\else.\null\fi\xspace}
\def\eg{\emph{e.g}\onedot}

\makeatother

\sethlcolor{red!20} 

\title{What to Test Next: Interpretable Coverage Gap Discovery in Driving VLMs}

\author{%
  Abhishek Aich$^{\dagger}$\thanks{Corresponding author: aaich001 \textit{at} ucr \textit{dot} edu}~, Sparsh Garg$^{\dagger}$, Vijay Kumar BG$^{\dagger}$, \\\textbf{Turgun Yusuf Kashgari}$^{\dagger}$, \textbf{Manmohan Chandraker}$^{\dagger, \ddagger}$   \\
  $^{\dagger}$NEC Laboratories, America, USA,
  $^{\ddagger}$University of California, San Diego, USA
}

\begin{document}

\maketitle

\begin{abstract}
Driving vision-language models (VLMs) must accurately understand scenes across diverse conditions defined by Operational Design Domains (ODDs), yet verification remains sparse: many slices are missing, making empirical failure rates unreliable. We propose \textsc{SliceScorer}, a deterministic scoring rule for missing-slice recommendation that combines (i) an exposure-based coverage prior to prioritize rare, under-tested regions, and (ii) a neighbor-failure prior that propagates risk from similar tested conditions. \textsc{SliceScorer} is deliberately simple - interpretable, auditable, and conservative - properties essential for safety-critical validation. For stress testing beyond the declared ODD, we embed \textsc{SliceScorer} within \textsc{SliceNav}, an LLM-orchestrated verification pipeline where the model interprets developer queries to select relevant operators (triage, scoring, acquisition, evaluation) and vocabulary extensions, composing verification workflows while keeping all scoring deterministic and auditable. Experiments on three driving VLMs (WiseAD, DriveMM, Cosmos-Reason2-2B) show that \textsc{SliceNav} surfaces high-risk coverage gaps more effectively than prior slice-discovery methods while maintaining diverse recommendations across the condition space. Ablations confirm both scoring components contribute, and qualitative analysis demonstrates end-to-end workflows from developer query to targeted evaluation. 

\end{abstract}

\section{Introduction}

Driving vision-language models (VLMs) must work reliably across diverse conditions - foggy nights, construction zones, dense traffic, unusual road layouts, \textit{etc.} - yet testing every scenario is infeasible. Standard benchmarks cover common cases well, but rare conditions where failures matter most remain undertested. When validation budgets are limited, developers need principled guidance: \emph{what should we test next?} This question is not addressed by existing methods \cite{chung2018slicefinder,sagadeeva2021sliceline} . 

Performance gap-discovery methods like SliceFinder~\cite{chung2018slicefinder} and SliceLine~\cite{sagadeeva2021sliceline} identify failure patterns in \emph{already-evaluated} data - useful for diagnosis, but silent on where to look next. Operational Design Domains (ODD) taxonomies and coverage metrics~\cite{ISO34503:2023,ISO21448_2022,de2024coverage} describe \emph{where} a system should operate, but do not rank which untested condition to prioritize. Further, simply building an large language model (LLM) based agent would not provide a deterministic, auditable rule for selecting the next scenario to evaluate - precisely what safety-critical verification requires (see \Cref{fig:teaser}). Given these important caveats, we pose an important problem in this paper: 
\begin{figure*}[!t]
  \centering
  \includegraphics[width=\linewidth]{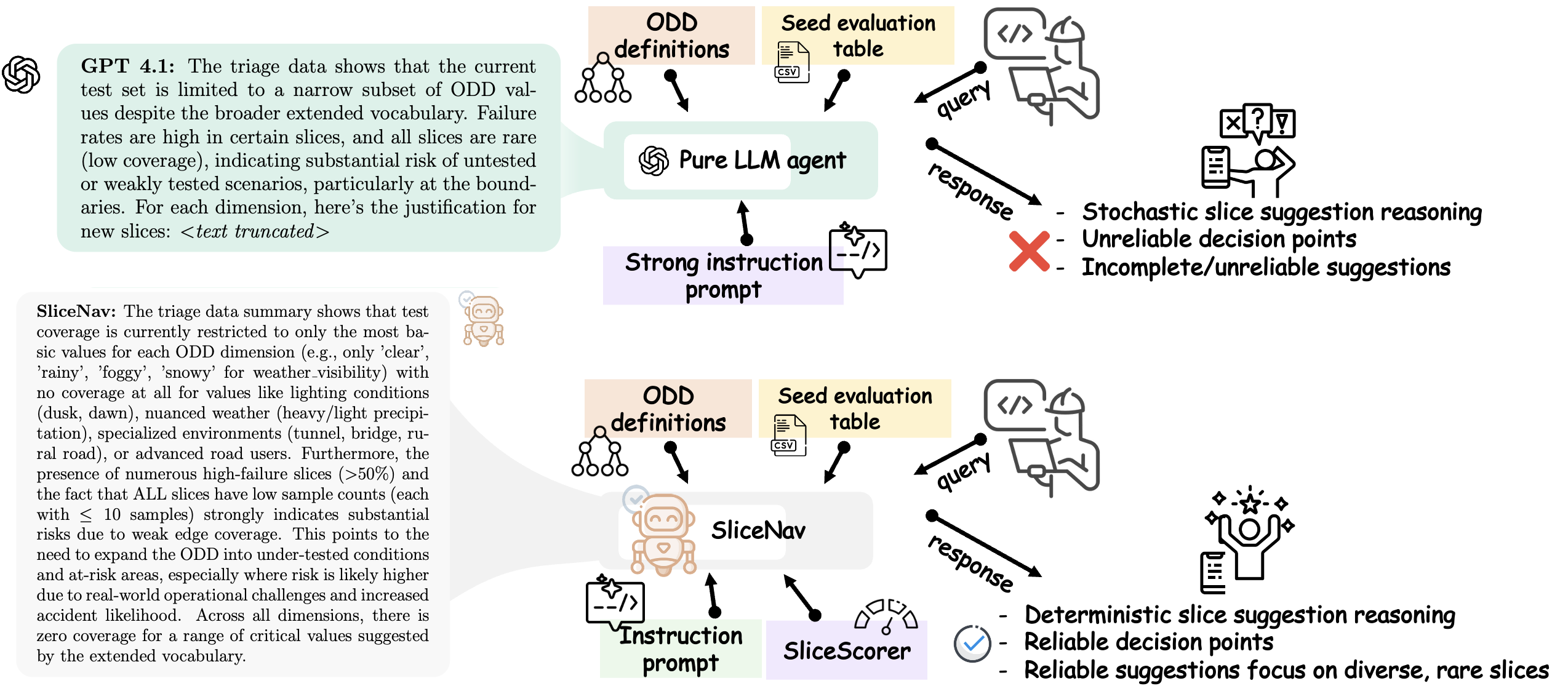}
  \caption{
\textbf{From stochastic prompting to structured slice reasoning.}
\emph{Top:} A pure LLM agent, even when provided with ODD definitions, seed evaluation tables, and strong instruction prompts, produces stochastic slice suggestions with unreliable decision boundaries and incomplete coverage. 
\emph{Bottom:} \textsc{SliceNav} augments the LLM with structured instructions and a dedicated \textsc{SliceScorer}, enabling deterministic slice suggestion reasoning, reliable decision points, and focused discovery of diverse, rare, and high-risk slices. This structured integration leads to more systematic, interpretable, and dependable ODD coverage expansion.
}
  \label{fig:teaser}
\end{figure*}
\begin{quote}
\textit{Given a validation set with partial coverage, how do we prioritize \textbf{missing} scenarios to test next - both within and beyond the declared operating domain - in a way that is traceable and auditable?}
\end{quote}

To this end, we introduce \textsc{SliceNav}, an agentic verification framework that recommends which untested driving scenarios to evaluate and orchestrates operator calls based on user query. The core insight is simple: rank missing scenarios by combining \textbf{rarity} - how undertested a condition is - with \textbf{neighbor-failure similarity} - how close it is to conditions where the model already fails. This scoring is  implemented in our proposed \textsc{SliceScorer}. An LLM then uses this scenario scoring mechanism for both inside-ODD and outside-ODD related developer queries to build an overall fully deterministic and auditable framework.

We evaluate our framework on three driving VLMs: WiseAD \cite{zhang2024wisead}, DriveMM \cite{huang2024drivemm}, and Cosmos-Reason2-2B (Cosmos) \cite{nvidia_cosmos_reason2_docs_2026}. and measure two complementary objectives: \emph{risk capture} (do top-ranked scenarios reveal actual failures?) and \emph{diversity} (do recommendations span the condition space rather than clustering in one region?). Experiments on three driving VLMs show that \textsc{SliceScorer} recommends a broader range of missing driving conditions than baselines \cite{chung2018slicefinder,sagadeeva2021sliceline} across the evaluated testing budgets, while also identifying higher-risk missing conditions more effectively at larger test-campaign budgets. At $K=5000$, \textsc{SliceScorer} reduces the average missed failure risk by 65\% on DriveMM and 71\% on Cosmos (compared to SliceFinder \cite{chung2018slicefinder}), while producing the highly diverse recommendation lists. Qualitative runs further demonstrate end-to-end workflows - from developer query through ODD extension, data acquisition, and VLM evaluation - with all scoring remaining deterministic. We summarize our contributions as follows:
\begin{enumerate}[leftmargin=*,itemsep=2pt]
    \item We formulate \textbf{missing-scenario recommendation} for driving VLM verification: given partial test coverage, rank untested conditions by predicted failure risk.
    \item We propose \textsc{SliceScorer}, a deterministic scoring rule combining rarity and neighbor-failure priors - interpretable, auditable, and simple by design.
    \item We present \textsc{SliceNav}, which embeds this scorer in an LLM-orchestrated workflow for flexible, query-driven verification while preserving deterministic ranking.
\end{enumerate}

\section{Related Works}
\label{sec:related}

\paragraph{Failure subpopulations discovery methods.} Slice discovery aims to diagnose model failures by identifying human-interpretable subpopulations (``slices'') where performance deviates significantly from the global average, typically expressed as conjunctions of attribute predicates. Early work such as SliceFinder~\cite{chung2018slicefinder} introduced heuristic lattice exploration and statistical controls to surface problematic tabular subgroups, while SliceLine~\cite{sagadeeva2021sliceline} improved scalability through aggressive pruning and a linear-algebraic formulation for efficient top-$K$ conjunctive slice enumeration. Subsequent systems extend this paradigm toward large-scale and interactive settings: AutoSlicer~\cite{autoslicer2022} emphasizes distributed computation and hypothesis testing for production-scale auditing, and Divisi~\cite{divisi2025} adopts sampling and constrained (e.g., beam-style) search for fast approximate discovery. Closely related are pattern-mining approaches such as DivExplorer~\cite{divexplorer2021}, which frame slice discovery as mining frequent itemsets with high metric divergence, integrating support thresholds and redundancy control. In contrast, representation-driven methods for unstructured domains \cite{domino2022} discover error-prone regions in embedding space, often sacrificing the direct interpretability of explicit predicate-defined slices. Our work follows the SliceLine-style paradigm~\cite{sagadeeva2021sliceline}, focusing on scalable, interpretable rule-based slice discovery over large candidate spaces to enable transparent and actionable model development.

\paragraph{Importance of evaluation reliability.} Recent multimodal evaluation pipelines increasingly use LLMs for scalable scoring of open-ended outputs and structured reporting, but such evaluators must be audited for bias and reliability~\cite{zheng2023mtbench,dubois2024lengthcontrolledalpacaeval,yu2023mmvet}. Separately, interpretability research emphasizes that explanations should be evaluated for faithfulness and practical utility rather than plausibility alone~\cite{doshi2017rigorous,jacovi2020faithful}. Motivated by these concerns, we implement our method as an agentic controller that orchestrates structured verification operators. 

\section{Proposed Method}
\label{sec:method}


\begin{figure}[t]
\centering
\includegraphics[width=\columnwidth]{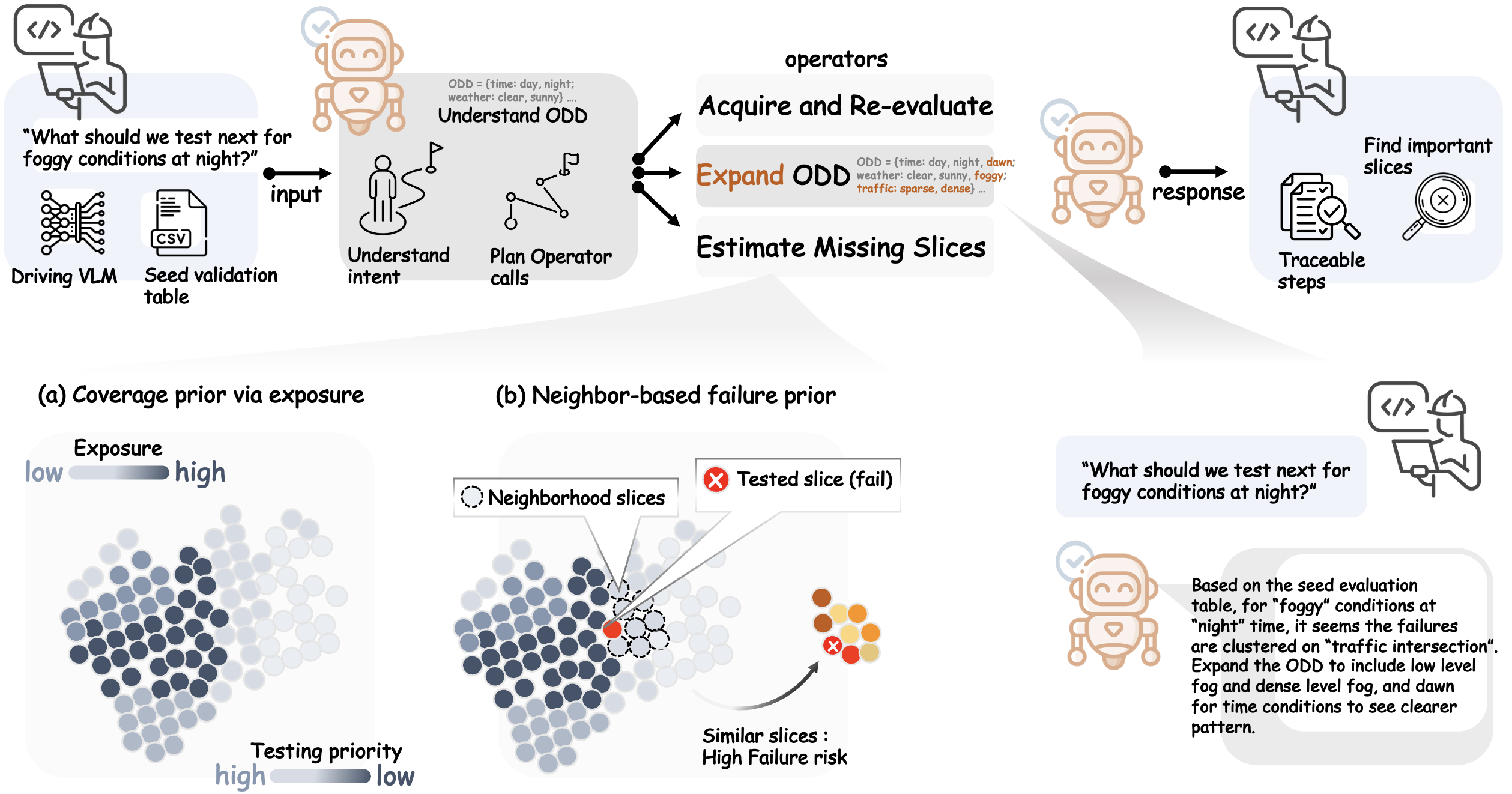}

\caption{\textbf{Framework overview.}
The agent takes a natural-language verification goal plus evaluation context and ODD specification, then
orchestrates failure triage, in-ODD gap discovery, and controlled out-of-ODD stress proposal via a query-conditioned
ODD expansion. It can also close the loop by acquiring data, and re-evaluating the VLM.}
\label{fig:main_fig}
\end{figure}

\subsection{Terminology}
\label{sec:terminology}

We first establish key terms used throughout this section. An \textbf{Operational Design Domain (ODD)} specifies the conditions under which an autonomous system is designed to operate; we instantiate the ODD as a set of condition dimensions (weather, time, road type, etc.) with discrete value sets. A \textbf{slice} is a specific combination of values across all ODD dimensions - for example, \emph{(foggy, night, highway, wet, dense traffic, intersection, hazard assessment)} - representing a distinct operating scenario. The \textbf{seed validation table} records VLM performance on evaluated slices from a benchmark dataset; each sample is human-verified and slices must meet a minimum sample threshold to ensure statistically meaningful measurements. In practice, this table is sparse - many ODD slices have zero samples - motivating the need for systematic gap identification. An \textbf{observed slice} has at least one sample in the seed validation table; a \textbf{missing slice} has zero samples, meaning VLM behavior under those conditions is unknown. A \textbf{coverage gap} refers to a missing slice - an ODD condition combination absent from the seed table. \textbf{Exposure} estimates how likely a slice is to appear in the validation data based on marginal frequencies; low-exposure slices are rare and under-tested. Finally, \textbf{next-test recommendation} is the task of selecting which missing slices to evaluate given a limited testing budget; \textsc{SliceNav} outputs a ranked list where higher-ranked slices are more likely to reveal failures or close critical coverage gaps.

\subsection{Problem Definition}
\label{sec:problem_definition}

\paragraph{ODD taxonomy and slice space.}
Let $K$ denote the number of ODD condition dimensions (\eg, weather visibility, sky conditions, time of day). Each dimension $k \in \{1,\dots,K\}$ has a finite in-ODD value set $V_k$. A slice is formally a complete assignment:
\begin{equation}
s = (s_1, \dots, s_K), \quad s_k \in V_k,
\end{equation}
and the in-ODD slice universe is the Cartesian product $\mathcal{S}_{\text{ODD}} = \prod_{k=1}^{K} V_k$. For each slice $s$, the seed table records the sample count $n(s)$ and aggregated performance metrics. We use LLM-judge scores uniformly across question types (free-form QA, multiple-choice, captioning), averaged across all samples for slice $s$. 

\paragraph{Observed and missing slices.}
The set of observed slices (those with evaluation support) is:
\begin{equation}
\mathcal{S}_{\text{obs}} = \{s \in \mathcal{S}_{\text{ODD}} : n(s) > 0\}.
\end{equation}
The missing slices are $\mathcal{S}_{\text{miss}} = \mathcal{S}_{\text{ODD}} \setminus \mathcal{S}_{\text{obs}}$. Our primary goal is to rank $s \in \mathcal{S}_{\text{miss}}$ so that high-risk coverage gaps are prioritized for testing.

\paragraph{User intent via query.}
A developer may provide a natural-language query $q$ describing a focus (\eg, ``foggy night hazards'' or ``construction zone failures'').

\subsection{Overview of the Workflow}
\label{sec:workflow}
\textsc{SliceNav} is an agentic VLM ODD verification framework that reframes slice analysis as a \emph{test selection strategy}: given a seed validation table, it recommends which untested condition slices should be evaluated next and keeps each recommendation traceable to observed evidence. It combines \emph{(i)}~a deterministic slice-prioritization model (\textsc{SliceScorer}), which scores untested slices using exposure rarity and observed failure evidence, with \emph{(ii)}~an agentic orchestration layer, where an LLM interprets developer queries, selects verification operators, and composes multi-step workflows. This treats ODD verification as a sequential decision problem over a combinatorial condition space, addressing a gap left by existing slice analysis tools~\cite{chung2018slicefinder,sagadeeva2021sliceline,eyuboglu2022domino}, which diagnose failures within already evaluated datasets but do not guide where evaluation effort should go next. Figure~\ref{fig:main_fig} summarizes the end-to-end workflow. Given the inputs above, \textsc{SliceNav} executes (all or a subset of) the following operators depending on the user query:
\begin{enumerate}[leftmargin=*,itemsep=2pt]
    \item \textbf{Triage:} Summarize where the VLM fails among observed slices.
    \item \textbf{In-ODD scoring:} Enumerate missing in-ODD slices and rank them using \textsc{SliceScorer}.
    \item \textbf{Out-of-ODD scoring:} Extend the ODD vocabulary and rank newly reachable missing slices.
    \item \textbf{Acquisition:} Retrieve matching samples for recommended slices.
    \item \textbf{Evaluation:} Run the VLM on acquired samples and update the validation table.
\end{enumerate}

The LLM translates developer intent into operator sequences and produces human-readable responses. This separation ensures that critical risk estimation remains deterministic (via \textsc{SliceScorer}) while enabling flexible interaction with verification workflows. A run of \textsc{SliceNav} produces a natural-language report summarizing failures, coverage gaps, and recommended next tests.

\subsection{Deterministic Slice Scoring with \textsc{SliceScorer}}
\label{sec:slicescorer}

We now define \textsc{SliceScorer}, the function used to rank missing slices $s \in \mathcal{S}_{\text{miss}}$. The scorer is fully determined by the seed validation table, making recommendations reproducible and auditable.

\paragraph{Design rationale.}
Unlike diagnostic methods like SliceFinder~\cite{chung2018slicefinder} and SliceLine~\cite{sagadeeva2021sliceline} that search for \emph{observed} subpopulations with high error, our goal is \emph{next-test selection}: prioritizing \emph{missing} slices. This requires: (i)~a notion of exposure rarity to identify under-tested regions, and (ii)~a way to extrapolate risk from observed evidence. We present \textsc{SliceScorer} $\phi$, a multiplicative combination of rarity and neighbor-failure priors:
\begin{equation}
\phi(s) = w_{\text{rare}}(s) \cdot w_{\text{nb}}(s), \qquad s \in \mathcal{S}_{\text{miss}}.
\label{eq:phi}
\end{equation}
Sorting missing slices by $\phi(s)$ yields a prioritized next-test list that prefers (i)~under-tested regions and (ii)~candidates that are similar to observed high-error slices. This decomposition yields an auditable rationale for each recommendation, essential when triggering costly or safety-critical testing. 

\paragraph{Rarity prior via exposure.}
Not every missing slice is equally worth testing. We estimate each slice's exposure and prioritize low-exposure gaps. Let $N = \sum_{s \in \mathcal{S}_{\text{obs}}} n(s)$ be the total evaluated support. For each dimension $k$ and each value $v$ that appears in the seed table, the marginal support is $n_k(v) = \sum_{s \in \mathcal{S}_{\text{obs}} : s_k = v} n(s)$. We apply Laplace smoothing to handle values with zero samples gracefully; without it, a single zero marginal would collapse the entire exposure to zero. The smoothed marginal probability (with smoothing parameter $\alpha > 0$) is:
\begin{equation}
P_k(v) = \frac{n_k(v) + \alpha}{N + \alpha |V_k|},
\label{eq:P}
\end{equation}
for values $v \in V_k$. We approximate joint exposure by a product of per-dimension probabilities, assuming dimension independence for tractability since computing true joint probabilities is infeasible with sparse data:
\begin{equation}
E(s) = \left( \prod_{k=1}^{K} P_k(s_k) \right)^{\tau},
\label{eq:E}
\end{equation}
where $\tau \in (0,1]$ reduces dynamic range so exposures remain comparable as $K$ increases. Finally, we convert exposure to a rarity-based priority by subtracting from 1 (so rare slices score high) and raising to a power $\gamma$ that controls how aggressively we prioritize rare over common slices:
\begin{equation}
w_{\text{rare}}(s) = (1 - E(s))^{\gamma}.
\label{eq:wrare}
\end{equation}
Note that our goal is to provide a deterministic and auditable prioritization rule that remains usable when seed coverage is sparse. The independence approximation is therefore used as a conservative coverage prior, and not as a  claim about the true data-generating distribution.

\paragraph{Neighbor-based failure prior.}
Rarity alone does not indicate \emph{which} missing slices are risky. We propagate failure evidence from \emph{observed} slices that are semantically close to the candidate. We use embedding similarity rather than exact dimension matches because it captures that related concepts (e.g., ``heavy\_fog'' and ``fog'') should share risk signals even if textually different. Each slice embedded with a text encoder; let $\text{sim}(s,s') \in [0,1]$ denote non-negative clipped cosine similarity between slice embeddings. The neighbor set $\mathcal{N}(s) \subseteq \mathcal{S}_{\text{obs}}$ contains the top-$m$ observed slices by similarity, restricted to those with $\text{sim}(s,s') \geq \theta$; if no neighbor exceeds this threshold, $\mathcal{N}(s) = \emptyset$. For an observed slice $s' \in \mathcal{S}_{\text{obs}}$, let $\epsilon(s') = 1 - \bar{y}(s') \in [0,1]$ denote the error rate, where $\bar{y}(s')$ is the mean LLM-judge score for slice $s'$. We define the neighbor-based error estimate:
\begin{equation}
\hat{\epsilon}(s) = \min\!\left(
\max_{s' \in \mathcal{N}(s)} \text{sim}(s,s') \cdot \epsilon(s'),
\; \kappa \right),
\label{eq:eps}
\end{equation}
where we take the max because we want the strongest risk signal. We multiply by similarity so closer neighbors provide more reliable evidence; and we cap with $\kappa \in (0,1]$ to prevent a single extreme neighbor from dominating. When $\mathcal{N}(s) = \emptyset$, we set $\hat{\epsilon}(s)=0$. The neighbor weight is:
\begin{equation}
w_{\text{nb}}(s) = 1 + \hat{\epsilon}(s).
\label{eq:wnb}
\end{equation}
The additive 1 ensures slices without neighbor evidence still receive a baseline score; the multiplicative form in $\phi(s)$ then boosts slices near observed failures proportionally, bounded by $(1+\kappa)$.

\subsection{\textsc{SliceNav}: Agentic Orchestration}
\label{sec:agentic}

While \textsc{SliceScorer} provides deterministic slice ranking, \textsc{SliceNav}'s agentic layer enables flexible, intent-driven verification workflows. The LLM makes two key decisions: \emph{operator selection} and \emph{outside-ODD extension}.

\paragraph{Operator selection.}
Given a developer's natural-language query, the LLM interprets the intent and selects which operators to invoke. For example, the query ``Where does the model fail on foggy nights?'' triggers triage followed by in-ODD scoring with a focus on foggy-night slices. The query ``Stress test beyond current ODD for construction scenarios'' triggers vocabulary extension and out-of-ODD scoring. This dispatch mechanism allows developers to interact with \textsc{SliceNav} conversationally while the underlying scoring remains deterministic.

\paragraph{Extension outside-ODD.}
Beyond identifying coverage gaps \emph{within} the declared ODD, we support stress-test recommendations \emph{outside} the base domain. We maintain an extended vocabulary of plausible out-of-ODD values (\eg, \emph{fog\_dense}, \emph{tunnel}, \emph{glare}). Given query $q$, the LLM selects relevant extensions, yielding expanded value sets $V_k^{+}(q) \supseteq V_k$ per dimension. The expanded slice space is:
\begin{equation}
\mathcal{S}^{+}(q) = \prod_{k=1}^{K} V_k^{+}(q).
\label{eq:expand}
\end{equation}
Missing slices in this expanded space are $\mathcal{S}_{\text{miss}}^{+}(q) = \mathcal{S}^{+}(q) \setminus \mathcal{S}_{\text{obs}}$. To score these out-of-ODD candidates, we adapt \textsc{SliceScorer} to handle novel vocabulary. Since out-of-vocabulary values $v \notin V_k$ have no marginal counts, we estimate their probability by matching to the nearest observed value: $P_k(v) = P_k(v^*) \cdot \text{sim}(v, v^*) \cdot \beta$, where $v^* \in V_k$ is the nearest observed value and $\text{sim}(v, v^*)$ denotes cosine similarity between their text embeddings, and $\beta \in (0,1]$ discounts confidence in the extrapolated estimate. The same $\phi(s)$ formula (\Cref{eq:phi}--\Cref{eq:wnb}) then applies to rank out-of-ODD candidates.

\paragraph{Workflow composition.}
Given a query, \textsc{SliceNav} chains operators into a verification workflow. \textsc{SliceScorer} keeps scoring deterministic, while the LLM handles intent, sequencing, and readable output - preserving reproducibility and auditability with natural-language interaction.

\section{Experiments}
\label{sec:experiments}

\paragraph{Baselines and VLMs.} Within-ODD baselines are SliceLine~\cite{sagadeeva2021sliceline} and SliceFinder~\cite{chung2018slicefinder}; since these methods were originally designed for discovering high-error \emph{observed} subpopulations rather than ranking \emph{missing} slices, we adapt them as described in Appendix~\ref{sec:baseline_adaptation}. For outside-ODD testing, we compare baselines include embedding-based k-Nearest neighbor (kNN) with $k=1$ retrieval and random selection. We use driving VLMs WiseAD \cite{zhang2024wisead}, DriveMM \cite{huang2024drivemm}, and Cosmos-Reason2-2B \cite{nvidia_cosmos_reason2_docs_2026}. 

\paragraph{Oracle ranking methodology.}
To evaluate slice recommendation quality, we construct an oracle slice table from a large, independently collected dataset (see Appendix \ref{sec:dataset_collection}, Table \ref{tab:oracle_set}) that is disjoint from the seed validation set (see Appendix \ref{sec:dataset_collection}, Table \ref{tab:sampled_set}). For each slice $s$, we evaluate the VLM on all corresponding oracle examples and compute per-example correctness using an LLM-based judge. Let $y_i \in [0,1]$ denote the judge score for oracle example $i$ in slice $s$. We aggregate these scores into the slice-level oracle estimate
\begin{equation}
\bar{y}_{\mathrm{oracle}}(s)
=
\frac{1}{N_{\mathrm{oracle}}(s)}
\sum_{i=1}^{N_{\mathrm{oracle}}(s)} y_i ,
\end{equation}
where $N_{\mathrm{oracle}}(s)$ is the number of oracle examples in slice $s$. Matching the notation in Section~3.4, we define the oracle slice failure risk as
$\epsilon_{\mathrm{oracle}}(s)=1-\bar{y}_{\mathrm{oracle}}(s)$. The oracle ranking is obtained by sorting slices in descending order of $\epsilon_{\mathrm{oracle}}(s)$, which serves as a high-confidence approximation of slice-level failure risk. To build the judge and prompt, we reused VLMEvalKit~\cite{duan2024vlmevalkit} functions for consistency; prompts and interface details are provided in Appendix~\ref{sec:llm-judge}.

\paragraph{Evaluation metrics.}
We assess next-test recommendations on two complementary objectives: failure discovery and diversity. We report both metrics together because the goal is to identify high-risk missing slices while also covering diverse operating conditions. Regret@$K$ measures how much oracle failure risk is missed relative to the best possible top-$K$ recommendation list:
\begin{equation}
\mathrm{Regret@}K
=
\frac{1}{K}
\left(
\sum_{i=1}^{K} \epsilon^{\star}_{i}
-
\sum_{i=1}^{K} \epsilon_i
\right),
\end{equation}
where $\epsilon^{\star}_{1} \geq \cdots \geq \epsilon^{\star}_{K}$ are the oracle failure risks of the top-$K$ slices under the oracle ranking, and $\epsilon_i$ is the oracle failure risk of the $i$-th slice returned by the evaluated method. Lower Regret indicates faster failure discovery. We also report ILD@$K$ (intra-list diversity), defined as the mean pairwise label disagreement across dimensions in the top-$K$ recommendation list~\cite{ziegler2005improving}; higher ILD indicates broader coverage of the condition space.

\paragraph{Ground-truth data construction.} The full ODD slice space forms a large Cartesian product over all dimensions (3,840 in-ODD; 91M+ outside-ODD), making exhaustive coverage infeasible. To build a diverse evaluation pool, we aggregated videos and images from multiple datasets~\cite{Shounak2024-rc, malla2023drama,corbiere2025drivingvqa,marcu2024lingoqa,xai_realworldqa_2024,xu2021sutd,Zhou2025-fe, pitropov2021canadian, peng2023pesotif, kenk2020dawn, alibeigi2023zenseact,sheeny2021radiate, gupta2024video, marathe2023wedge,zendel2018wilddash} focused on diverse weather (see Appendix~\ref{sec:dataset_collection}). We annotated each with in-ODD and extended-ODD attributes using Qwen3-30B~\cite{qwen3}. For QA task type, we categorized existing QA pairs using GPT-5.2\cite{OpenAI2025GPT52}, using them as few-shot demonstrations to assign pseudo ground-truth labels. We had four domain experts independently verify our predicted scene attributes and QA metadata on 100 randomly drawn examples; inter-rater agreement (Fleiss \cite{fleiss1971}) was $0.2395$ with the mean observed pairwise agreement as 0.8263, across all judgments. For acquisition-based workflows in Analysis~\ref{sec:analysis_3}, \textsc{SliceNav} queries a retrieval datastore of driving footage to obtain candidate samples for recommended slices; we describe the datastore sources, scale, and retrieval setup in Appendix~\ref{sec:datastore_retrieval_slicenav}.

\paragraph{ODD taxonomy.}
Our base ODD taxonomy is grounded in ISO~34503, ISO~21448 (SOTIF), and SAE~J3016 (see Appendix~\ref{sec:odd_standards_mapping} for detailed standards mapping). It uses eight structured dimensions describing environmental conditions, road context, traffic state, and task type. Specifically, \texttt{weather\_visibility} captures coarse weather (\texttt{snowy}, \texttt{foggy}, \texttt{clear}, \texttt{rainy}); \texttt{sky\_conditions} distinguishes cloud cover (\texttt{cloudy}, \texttt{clear}); and \texttt{time} models illumination (\texttt{day}, \texttt{night}). The driving context is described by \texttt{environment} (e.g., \texttt{city downtown}, \texttt{city suburbs}, \texttt{construction}, \texttt{highway}), \texttt{road\_condition} (e.g., \texttt{dry}, \texttt{wet}, \texttt{snow}), \texttt{road\_type} (\texttt{intersection}, \texttt{non-intersection}), and \texttt{traffic\_type} (\texttt{sparse}, \texttt{dense}). Finally, \texttt{question\_category} specifies the QA task (e.g., scene observation, hazard assessment, future event prediction).

To enable controlled stress testing beyond the declared domain, we define an extended ODD vocabulary that refines and expands each dimension. Weather is decomposed into finer-grained intensities (e.g., \texttt{rain\_light}, \texttt{rain\_heavy}, \texttt{fog\_dense}, \texttt{snow\_heavy}) and additional phenomena (e.g., \texttt{dust}, \texttt{smoke}, \texttt{haze}). Sky and illumination are extended with conditions such as \texttt{partly\_cloudy}, \texttt{overcast}, \texttt{sun\_glare}, \texttt{dawn}, and \texttt{dusk}. Environmental and road context dimensions are broadened to include specific scene types (e.g., \texttt{industrial\_area}, \texttt{parking\_lot}, \texttt{tunnel}, \texttt{bridge}, \texttt{rural\_road}) and more detailed road structures (e.g., \texttt{roundabout}, \texttt{merge\_ramp}, \texttt{intersection\_signalized}). Traffic is refined into finer density regimes (e.g., \texttt{moderate}, \texttt{stop\_and\_go}, \texttt{traffic\_jam}), and the previously empty \texttt{road\_users\_present} dimension is introduced to explicitly model interacting agents (e.g., \texttt{pedestrians}, \texttt{cyclists}, \texttt{mixed\_vulnerable\_users}). The extended ODD thus provides a structured superset of the base domain, enabling systematic exploration out-of-domain conditions. Note that dimensions that are not active in the base ODD, such as \texttt{road\_users\_present}, are assigned a default base value unspecified; outside-ODD expansion adds additional values to the same dimension rather than introducing a new tuple axis.

\paragraph{Experimental protocol.} We define the exact experimental protocol for Analysis~1--4 in Appendix~\ref{sec:experimental_protocols}, including inputs, outputs, and how we adapt baselines for fair comparison. Further, \textsc{SliceScorer} uses hyperparameters: smoothing $\alpha = 1.0$ (\Cref{eq:P}), dynamic range $\tau = 0.6$ (\Cref{eq:E}), rarity emphasis $\gamma = 2.0$ (\Cref{eq:wrare}), neighbor cap $\kappa = 0.8$ (\Cref{eq:eps}), neighbor count $m = 5$, similarity threshold $\theta = 0.3$ for scoring, and additionally decay $\beta = 0.8$ for out-of-ODD vocabulary extension. We embed slice tuples with the text encoder model \texttt{all-MiniLM-L6-v2} \cite{reimers2019sentence} using cosine similarity. We avoided dataset-specific sweeps to keep comparisons reproducible and fair across VLMs and analyses.

\subsection{Analysis 1: In-ODD missing-slice ranking}
\begin{table*}[t]
\centering
\caption{\textbf{In-ODD analysis.} Lower Regret indicates faster failure discovery and higher ILD indicates broader coverage. \textsc{SliceScorer} generally improves diversity and achieves the strongest Regret at larger budgets, reflecting a coverage-risk tradeoff than baselines. SliceLine (below separator) is shown as a coarse-reference baseline on observed errors, not a full-tuple missing-slice ranker.}
\label{tab:in_odd_vlm_comparison}
\small
\begin{subtable}[t]{0.31\textwidth}
  \centering
  \caption{WiseAD}
  \label{subtab:in_odd_wisead}
  \setlength{\tabcolsep}{4pt}%
\resizebox{\linewidth}{!}{%
\begin{tabular}{@{}llcc@{}}
\hline
$K$ & Method & Regret@$K$ $\downarrow$ & ILD@$K$ $\uparrow$ \\
\hline
500 & \textsc{SliceScorer} & 0.5618 & 0.5788 \\
500 & SliceFinder & 0.5103 & 0.5321 \\
\cline{1-4}
500 & \textcolor{gray}{SliceLine} & \textcolor{gray}{0.5151} & \textcolor{gray}{0.5270} \\
\hline
1000 & \textsc{SliceScorer} & 0.3851 & 0.5980 \\
1000 & SliceFinder & 0.3460 & 0.5724 \\
\cline{1-4}
1000 & \textcolor{gray}{SliceLine} & \textcolor{gray}{0.3357} & \textcolor{gray}{0.5676} \\
\hline
5000 & \textsc{SliceScorer} & 0.0258 & 0.6203 \\
5000 & SliceFinder & 0.0764 & 0.5724 \\
\cline{1-4}
5000 & \textcolor{gray}{SliceLine} & \textcolor{gray}{0.0302} & \textcolor{gray}{0.6200} \\
\hline
\end{tabular}%
}

\end{subtable}
\hfill
\begin{subtable}[t]{0.31\textwidth}
  \centering
  \caption{DriveMM}
  \label{subtab:in_odd_drivemm}
  \setlength{\tabcolsep}{4pt}%
\resizebox{\linewidth}{!}{%
\begin{tabular}{@{}llcc@{}}
\hline
$K$ & Method & Regret@$K$ $\downarrow$ & ILD@$K$ $\uparrow$ \\
\hline
500 & \textsc{SliceScorer} & 0.3493 & 0.5535 \\
500 & SliceFinder & 0.3193 & 0.5220 \\
\cline{1-4}
500 & \textcolor{gray}{SliceLine} & \textcolor{gray}{0.2882} & \textcolor{gray}{0.5208} \\
\hline
1000 & \textsc{SliceScorer} & 0.2105 & 0.5741 \\
1000 & SliceFinder & 0.1809 & 0.5693 \\
\cline{1-4}
1000 & \textcolor{gray}{SliceLine} & \textcolor{gray}{0.1536} & \textcolor{gray}{0.5829} \\
\hline
5000 & \textsc{SliceScorer} & 0.0130 & 0.6204 \\
5000 & SliceFinder & 0.0369 & 0.5693 \\
\cline{1-4}
5000 & \textcolor{gray}{SliceLine} & \textcolor{gray}{0.0136} & \textcolor{gray}{0.6204} \\
\hline
\end{tabular}%
}

\end{subtable}
\hfill
\begin{subtable}[t]{0.31\textwidth}
  \centering
  \caption{Cosmos}
  \label{subtab:in_odd_cosmos}
  \setlength{\tabcolsep}{4pt}%
\resizebox{\linewidth}{!}{%
\begin{tabular}{@{}llcc@{}}
\hline
$K$ & Method & Regret@$K$ $\downarrow$ & ILD@$K$ $\uparrow$ \\
\hline
500 & \textsc{SliceScorer} & 0.4847 & 0.5852 \\
500 & SliceFinder & 0.4419 & 0.5279 \\
\cline{1-4}
500 & \textcolor{gray}{SliceLine} & \textcolor{gray}{0.3918} & \textcolor{gray}{0.5341} \\
\hline
1000 & \textsc{SliceScorer} & 0.3114 & 0.5885 \\
1000 & SliceFinder & 0.2453 & 0.5718 \\
\cline{1-4}
1000 & \textcolor{gray}{SliceLine} & \textcolor{gray}{0.2523} & \textcolor{gray}{0.5804} \\
\hline
5000 & \textsc{SliceScorer} & 0.0154 & 0.6204 \\
5000 & SliceFinder & 0.0530 & 0.5718 \\
\cline{1-4}
5000 & \textcolor{gray}{SliceLine} & \textcolor{gray}{0.0186} & \textcolor{gray}{0.6201} \\
\hline
\end{tabular}%
}

\end{subtable}
\end{table*}

\textsc{SliceScorer} is designed to optimize a \emph{coverage--risk tradeoff}, not Regret alone. At small $K$, methods that concentrate around known high-error marginals can sometimes achieve lower Regret, but often at the cost of reduced diversity. Our objective is to recommend test suites that are both risky and broad, since coverage expansion is the verification goal. Table~\ref{tab:in_odd_vlm_comparison} supports this tradeoff. On DriveMM at $K{=}500$, \textsc{SliceScorer} has higher Regret than SliceFinder but achieves higher ILD, indicating broader coverage. At $K{=}5000$, \textsc{SliceScorer} obtains much lower Regret than SliceFinder while maintaining the strongest or tied-strongest ILD. Thus, \textsc{SliceScorer} is not optimized to greedily minimize Regret at the earliest budgets; it is designed for diverse, auditable, high-risk coverage expansion and becomes substantially stronger at deeper, practically relevant test-campaign budgets. Note that although $K{=}500$ or $1000$ may seem large, these budgets are sparse relative to the combinatorial ODD space: 3,840 in-ODD slices and over 91M outside-ODD combinations. Exhaustive coverage is therefore infeasible, motivating budgeted ranking for test-campaign construction rather than single-cluster failure mining.

\paragraph{Ablations.}
Across all datasets, we observe that a coverage-only (optimizing for rarity) variant underperforms on the joint objective, showing rarity alone is insufficient. The coverage-only variant, which optimizes rarity without neighbor-based failure propagation, underperforms on the joint objective. For example, at $K{=}500$, Regret worsens from $0.5583$ to $0.6265$ on WiseAD and from $0.4855$ to $0.5155$ on Cosmos relative to the neighbor-only variant. Diversity also drops substantially: ILD@$1000$ decreases from $0.5996$ to $0.5638$ on WiseAD and from $0.5912$ to $0.5638$ on Cosmos. These results indicate that neighbor-based failure propagation is important for early failure discovery, while the full scorer provides the most stable risk--coverage tradeoff.

\subsection{Analysis 2: Outside-ODD missing slice ranking}
\begin{table*}[t]
\centering
\caption{\textbf{Outside-ODD analysis.} \textsc{SliceScorer} is competitive with or slightly better than kNN on outside-ODD Regret, while providing an explicit rarity-risk decomposition. `Random' achieves high ILD by construction but does not prioritize failures.}
\label{tab:outside_odd_vlm_comparison}
\small
\begin{subtable}[t]{0.31\textwidth}
  \centering
  \caption{WiseAD}
  \label{subtab:out_odd_wisead}
  \setlength{\tabcolsep}{3pt}%
\resizebox{\linewidth}{!}{%
\begin{tabular}{@{}llcc@{}}
\hline
$K$ & Method & Regret@$K$ $\downarrow$ & ILD@$K$ $\uparrow$ \\
\hline
500 & \textsc{SliceScorer} & 0.9940 & 0.4453 \\
500 & Embedding kNN & 1.0000 & 0.4789 \\
500 & Random & 1.0000 & 0.8003 \\
\hline
1000 & \textsc{SliceScorer} & 0.9930 & 0.4806 \\
1000 & Embedding kNN & 0.9980 & 0.5037 \\
1000 & Random & 0.9990 & 0.8002 \\
\hline
5000 & \textsc{SliceScorer} & 0.3532 & 0.5560 \\
5000 & Embedding kNN & 0.3544 & 0.5776 \\
5000 & Random & 0.3560 & 0.8000 \\
\hline
\end{tabular}%
}

\end{subtable}
\hfill
\begin{subtable}[t]{0.31\textwidth}
  \centering
  \caption{DriveMM}
  \label{subtab:out_odd_drivemm}
  \setlength{\tabcolsep}{2pt}%
\resizebox{\linewidth}{!}{%
\begin{tabular}{@{}llcc@{}}
\hline
$K$ & Method & Regret@$K$ $\downarrow$ & ILD@$K$ $\uparrow$ \\
\hline
500 & \textsc{SliceScorer} & 0.9979 & 0.3979 \\
500 & Embedding kNN & 1.0000 & 0.4291 \\
500 & Random & 1.0000 & 0.8003 \\
\hline
1000 & \textsc{SliceScorer} & 0.7616 & 0.4421 \\
1000 & Embedding kNN & 0.7626 & 0.4631 \\
1000 & Random & 0.7626 & 0.8002 \\
\hline
5000 & \textsc{SliceScorer} & 0.1827 & 0.5405 \\
5000 & Embedding kNN & 0.1825 & 0.5554 \\
5000 & Random & 0.1830 & 0.8000\\
\hline
\end{tabular}%
}

\end{subtable}
\hfill
\begin{subtable}[t]{0.31\textwidth}
  \centering
  \caption{Cosmos}
  \label{subtab:out_odd_cosmos}
  \setlength{\tabcolsep}{2pt}%
\resizebox{\linewidth}{!}{%
\begin{tabular}{@{}llcc@{}}
\hline
$K$ & Method & Regret@$K$ $\downarrow$ & ILD@$K$ $\uparrow$ \\
\hline
500 & \textsc{SliceScorer} & 0.9743 & 0.5032 \\
500 & Embedding kNN & 0.9744 & 0.4812 \\
500 & Random & 0.9744 & 0.8003 \\
\hline
1000 & \textsc{SliceScorer} & 0.6608 & 0.5348 \\
1000 & Embedding kNN & 0.6599 & 0.5428 \\
1000 & Random & 0.6611 & 0.8002 \\
\hline
5000 & \textsc{SliceScorer} & 0.1507 & 0.6132 \\
5000 & Embedding kNN & 0.1508 & 0.6457 \\
5000 & Random & 0.1511 & 0.8000 \\
\hline
\end{tabular}%
}

\end{subtable}
\end{table*}

Outside the base ODD, we compare methods on the same expanded candidate pool (Table~\ref{tab:outside_odd_vlm_comparison}). Note that random selection achieves high ILD by construction (uniformly sampling the space is maximally diverse) but provides no failure prioritization - it serves as a diversity ceiling, not a practical baseline.

\textsc{SliceScorer} consistently achieves the lowest or tied-lowest Regret, discovering failures earlier than kNN and random. While random trivially maximizes diversity, \textsc{SliceScorer} maintains reasonable ILD (0.4--0.6) while improving risk capture. This is the key tradeoff: random finds nothing in particular across everywhere; \textsc{SliceScorer} finds failures across diverse conditions.

\subsection{Analysis 3: Agentic workflow examples with Cosmos  and LLM as GPT 5.2 (qualitative)}
\label{sec:analysis_3}
\begin{figure*}[t]
\centering
\footnotesize
\setlength{\tabcolsep}{3pt}
\renewcommand{\arraystretch}{1.12}
\begin{tabular}{p{0.04\linewidth}p{0.31\linewidth}p{0.23\linewidth}p{0.32\linewidth}}
\toprule
\textbf{Ex.} & \textbf{Developer query and tools} & \textbf{Vocabulary selected} & \textbf{Top recommendations and evidence} \\
\midrule
\textbf{A}
&
\textbf{Query:} ``What driving scenarios outside our current ODD should we prioritize for testing?''\newline
\textbf{Tools:} find missing in-ODD slices $\rightarrow$ extend ODD vocabulary $\rightarrow$ rank missing outside-ODD slices
&
\texttt{+dawn}, \texttt{+dusk}, \texttt{+puddles}, \texttt{+muddy}, \texttt{+icy}, \texttt{+light\_rain}, \texttt{+heavy\_rain}
&
\textbf{Top-2 slices:} (\texttt{clear}, \texttt{clear}, \texttt{dusk}, \texttt{city downtown}, \texttt{dry}, \texttt{intersection}, \texttt{sparse}); (\texttt{clear}, \texttt{clear}, \texttt{dusk}, \texttt{city downtown}, \texttt{wet}, \texttt{intersection}, \texttt{sparse}).
\\
& \multicolumn{3}{@{}p{0.84\linewidth}@{}}{\textbf{Rationale:} expands lighting and road-surface values adjacent to current high-failure rainy/intersection patterns.} \\
\midrule
\textbf{B}
&
\textbf{Query:} ``What tunnel and underpass scenarios need testing?''\newline
\textbf{Tools:} explicit vocabulary match $\rightarrow$ rank missing outside-ODD slices $\rightarrow$ summarize existing failures
&
\texttt{+tunnel}; no \texttt{underpass} slice surfaced in the top outside-ODD recommendations
&
\textbf{Top-2 slices:} (\texttt{foggy}, \texttt{clear}, \texttt{day}, \texttt{tunnel}, \texttt{dry}, \texttt{intersection}, \texttt{dense}); (\texttt{foggy}, \texttt{clear}, \texttt{day}, \texttt{tunnel}, \texttt{dry}, \texttt{non-intersection}, \texttt{dense}).
\\
& \multicolumn{3}{@{}p{0.84\linewidth}@{}}{\textbf{Rationale:} targets a user-specified deployment gap and varies road geometry after selecting the new environment value.} \\
\midrule
\textbf{C}
&
\textbf{Query:} ``Find untested night driving scenarios, generate synthetic test data, and run VLM evaluation.''\newline
\textbf{Tools:} rank missing in-ODD slices $\rightarrow$ retrieve images $\rightarrow$ run VLM evaluation $\rightarrow$ analyze by conditions
&
No vocabulary extension; the controller switches from slice discovery to acquisition and evaluation
&
\textbf{Top-2 slices:} (\texttt{rainy}, \texttt{cloudy}, \texttt{night}, \texttt{city suburbs}, \texttt{wet}, \texttt{intersection}, \texttt{sparse}, \texttt{hazard assessment}); (\texttt{rainy}, \texttt{cloudy}, \texttt{night}, \texttt{city suburbs}, \texttt{snow}, \texttt{intersection}, \texttt{sparse}, \texttt{hazard assessment}).
\\
& \multicolumn{3}{@{}p{0.84\linewidth}@{}}{\textbf{Evidence:} generated 10 synthetic samples; on the 8 night slices, MCQ accuracy was 0.0 across categories, with the worst slice also scoring 0.0 VQA and 0.0 judge.} \\
\bottomrule
\end{tabular}
\caption{\textbf{Agentic workflow examples for qualitative analysis.} Each row is one completed run from the 15-query qualitative sweep. Broad queries trigger vocabulary extension across multiple ODD dimensions, focused queries select targeted outside-ODD values, and acquisition queries chain slice discovery to data generation/retrieval and model evaluation (see rationales in Appendix~\ref{sec:qualitative_rationales}).}
\label{fig:agentic_workflow_examples}
\end{figure*} \Cref{fig:agentic_workflow_examples} reports three completed runs from the qualitative sweep, selected to cover broad outside-ODD exploration, focused outside-ODD intent, and a full acquisition/evaluation loop. The examples show that \textsc{SliceNav} is not only producing natural-language rationales: it selects different operator sequences from the user query, extends the ODD vocabulary when needed, and then passes the resulting candidate set to deterministic \textsc{SliceScorer} ranking. In the broad case, \textsc{SliceNav} uses the LLM controller and expands multiple dimensions (\eg, transitional lighting and road-surface states); in the focused case, it uses the query to add a targeted environment value (\texttt{tunnel}); in the acquisition case, it skips vocabulary extension and chains missing-slice discovery VLM evaluation. Appendix~\ref{sec:additional_qualitative_examples} provides additional completed examples covering generic acquisition, foggy-highway retrieval/evaluation, snow/ice coverage, dawn/dusk illumination, and construction-zone expansion.

\subsection{Analysis 4: Reliability and stochasticity of pure LLM framework (with Cosmos)}
To check the reliability and stochasticity of a pure LLM agent, we implemented a strong LLM agent baseline instructed to output missing-slice suggestions, and compared it with our tool-grounded pipeline across 15 queries (listed in the Appendix~\ref{sec:analysis4_queries}). We set GPT-5.2 as both frameworks' LLM. The baseline's prompts have been given in Appendix~\ref{sec:llm_choose_slices_prompts}. We repeated each query 10 times and recorded both slice-suggestion variance. For slice variance, we measure returned list-size variability (coefficient of variation over `number of slices requested') and pairwise set overlap (Jaccard \cite{jaccard1901}) across runs using full slice tuples. The pure LLM baseline is highly unstable: average pairwise Jaccard is only 0.027. This shows both count drift and near-disjoint slice sets across runs. In comparison, \textsc{SliceNav}  is far more stable: average pairwise Jaccard is 0.957. 

\section{Conclusions}
\label{sec:conclusion}
We present \textsc{SliceNav}, an agentic framework for systematic ODD verification of driving VLMs that turns sparse validation coverage into traceable recommendations for what to test next. The core component, \textsc{SliceScorer}, deterministically ranks missing full-slice scenarios by combining two complementary signals: exposure rarity, which prioritizes under-tested regions of the ODD, and failure evidence from nearby observed slices, which directs testing toward conditions likely to reveal model weaknesses. This design makes the resulting coverage-risk trade-offs interpretable and auditable under limited testing budgets. Around this scorer, \textsc{SliceNav} uses an agentic layer to map developer queries to composable verification operators, including failure triage, in-ODD gap discovery, out-of-ODD stress proposal, acquisition, and evaluation. Across quantitative studies and qualitative workflows, \textsc{SliceNav} supports diverse, risk-aware scenario selection for broad exploration, targeted testing, and closed-loop acquisition/evaluation. In  practical workflows, developers must decide which scenarios to evaluate next under limited testing budgets. \textsc{SliceNav} provides an interpretable and traceable mechanism for making this decision systematically.

\begin{ack}

We would like to thank Manyi Yao (University of California, Riverside), Mauricio Soroco (Simon Fraser University), and Sathvik Basavaraju (NEC Laboratories America), who, together with Sparsh Garg, helped validate the annotated examples used for method benchmarking.
\end{ack}

\bibliographystyle{unsrt}
\bibliography{main}

\newpage
\appendix

\section*{Supplementary Material}
\phantomsection
\label{sec:supplementary_material}
\startcontents[appendix]
\subsection*{Contents}
\printcontents[appendix]{}{1}{\setcounter{tocdepth}{2}}
\newpage

\section{Grounding Slice Dimensions in ODD Standards}
\label{sec:odd_standards_mapping}

Our slice space dimensions are derived from established ODD specification standards. \Cref{tab:odd_mapping} provides a detailed mapping of each slice dimension to specific clauses in ISO~34503:2023~\cite{ISO34503:2023}, ISO~21448:2022 (SOTIF)~\cite{ISO21448_2022}, and SAE~J3016~\cite{sae2014taxonomy}.

\paragraph{Standards Overview.}
ISO~34503:2023 defines a hierarchical ODD taxonomy with three top-level categories (Clause~8.2): \emph{scenery elements} (roads, junctions), \emph{environmental conditions} (weather, illumination), and \emph{dynamic elements} (traffic agents). ISO~21448 (SOTIF) identifies triggering conditions that can activate functional insufficiencies in perception systems, with Annex~B providing a 6-layer scenario factor model. SAE~J3016 defines ODD as operating conditions including ``environmental, geographical, and time-of-day restrictions, and/or the requisite presence or absence of certain traffic or roadway characteristics'' (Section~3.21).

\begin{table}[th!]
\caption{\textbf{Mapping of slice dimensions to ODD standards.} Clause references are aligned with ISO~34503:2023 taxonomy structure and ISO~21448 Annex~B layers. This table provides an alignment between our slice dimensions and representative ODD/scenario-factor concepts in ISO~34503, ISO~21448, and SAE~J3016; the mapping is intended to justify coverage of major ODD-relevant factors rather than to reproduce each standard's taxonomy verbatim.}
\label{tab:odd_mapping}
\small
\resizebox{\columnwidth}{!}{
\begin{tabular}{@{}llll@{}}
\toprule
\textbf{Slice Dimension} & \textbf{ISO 34503:2023} & \textbf{ISO 21448 (SOTIF)} & \textbf{SAE J3016 \S3.21} \\
\midrule
\texttt{weather visibility}
& Clause 10 (Environmental conditions: weather, visibility)
& SOTIF scenario factors: (Environmental conditions)
& ``environmental'' \\

\texttt{sky conditions}
& Clause 10 (Environmental conditions: illumination)
& SOTIF scenario factors: (Environmental conditions)
& ``environmental'' \\

\texttt{time}
& Clause 10 (Environmental conditions: illumination/day-night)
& SOTIF scenario factors: (Environmental conditions)
& ``time-of-day restrictions'' \\

\texttt{environment}
& Clause 9 (Scenery elements: operational area, road context)
& SOTIF scenario factors: (Road-level environment)
& ``geographical'' \\

\texttt{road condition}
& Clause 9 (Scenery elements: surface condition)
& SOTIF scenario factors: (Road-level environment)
& ``roadway characteristics'' \\

\texttt{road type}
& Clause 9 (Scenery elements: road type, junctions)
& SOTIF scenario factors: (Road-level environment)
& ``roadway characteristics'' \\

\texttt{traffic type}
& Clause 11 (Dynamic elements: traffic participants)
& SOTIF scenario factors: (Dynamic elements)
& ``traffic characteristics'' \\
\bottomrule
\end{tabular}}
\end{table}

\paragraph{Coverage and Scope.}
Our seven ODD-derived dimensions cover all three top-level ISO~34503 categories: environmental conditions (\texttt{weather\_visibility}, \texttt{sky\_conditions}, \texttt{time}), scenery elements (\texttt{environment}, \texttt{road\_condition}, \texttt{road\_type}), and dynamic elements (\texttt{traffic\_type}). The \texttt{question\_category} dimension is VLM-specific and captures the type of reasoning task (e.g., scene observation, hazard assessment, future prediction); it is not derived from ODD standards but is essential for evaluating VLM capabilities across different question modalities.

\paragraph{Extensibility.}
ISO~34503 Clause~8.1 explicitly allows stakeholders to extend the taxonomy with additional attributes when the base taxonomy does not adequately represent a specific operating environment. Our framework supports this extensibility: developers can add new dimensions or refine existing value sets as needed for their specific ODD verification requirements.

\section{Baseline Adaptation Details}
\label{sec:baseline_adaptation}

SliceLine~\cite{sagadeeva2021sliceline} and SliceFinder~\cite{chung2018slicefinder} were originally designed for a different task: discovering high-error \emph{observed} subpopulations (partial predicates with non-zero support). Our task is ranking \emph{missing} full slices (zero-support combinations) by predicted failure risk. This section documents how we adapt these methods for fair comparison.

\subsection{Task Mismatch}

\textbf{Original task:} Given a dataset with observed examples, find partial predicates (e.g., \texttt{weather\_visibility=foggy AND time=night}) that have high error rates. These predicates describe subpopulations \emph{present} in the data.

\textbf{Our task:} Given a sparse validation set, rank \emph{missing} full slices (complete assignments of all ODD dimensions with zero observed support) by predicted failure risk, to guide acquisition of new test cases.

Neither SliceLine nor SliceFinder directly addresses missing-slice ranking. We therefore adapt each method as described below.

\subsection{SliceLine Adaptation}

SliceLine uses a lattice-based search to find partial predicates (subsets of dimension-value constraints) with statistically significant error rates. We adapt it for missing-slice ranking via predicate expansion:

\begin{enumerate}[leftmargin=*]
    \item \textbf{Fit SliceLine} on the observed triage table, treating $\epsilon(s)$ as the error signal.
    \item \textbf{Extract predicates} from SliceLine's top-$k$ discovered slices, each with an associated error score.
    \item \textbf{Expand to full slices:} For each candidate full slice $s$, find all predicates whose constraints are satisfied by $s$. Assign $s$ the score of its best-matching (highest-scoring) predicate.
    \item \textbf{Rank by specificity then discovery order:} Sort matched slices first by predicate specificity (number of constrained dimensions, descending), then by predicate discovery rank (ascending), then by predicate score (descending), with slice identifier as the final deterministic tie-breaker.
\end{enumerate}

This expansion allows SliceLine's partial-predicate signal to propagate to full slices, including those with zero support.

\begin{figure}[t]
\centering
\caption{SliceLine Adaptation for Missing-Slice Ranking}
\label{alg:sliceline_adapt}
\begin{minipage}{0.97\linewidth}
\small
\begin{lstlisting}[
  mathescape=true,
  basicstyle=\scriptsize\ttfamily,
  columns=fullflexible,
  frame=single,
  breaklines=true,
  aboveskip=0.5em,
  belowskip=0.5em
]
Require: Triage table $T$ with columns $\{d_1, \ldots, d_K\}$ and error $e_i = 1 - y_i$
Require: Candidate full slices $\mathcal{S}$
Ensure: Ranked list of slices
1: Fit SliceLine on $T$ with error $e$; extract $\mathcal{P} = \{(C_j, \sigma_j)\}_{j=1}^{m}$  // $C_j$: constraints, $\sigma_j$: score
2: for each slice $s \in \mathcal{S}$ do
3:   $\mathcal{M}_s \gets \{(C_j, \sigma_j) \in \mathcal{P} : s \text{ satisfies } C_j\}$
4:   if $\mathcal{M}_s \neq \emptyset$ then
5:     $(C^*, \sigma^*) \gets \arg\max_{(C,\sigma) \in \mathcal{M}_s} \sigma$
6:     assign $s$: specificity $= |C^*|$, score $= \sigma^*$, rank $= j^*$
7: return slices sorted by (specificity desc, rank asc, score desc, slice id asc)
\end{lstlisting}
\end{minipage}
\end{figure}

\subsection{SliceFinder-Inspired Marginal-Lift Model}

The original SliceFinder uses decision-tree-based partitioning to discover interpretable high-error subgroups. Since decision trees require observed support, we adapt the core intuition---that slice error can be decomposed into interpretable factors---using an \emph{additive marginal-lift model}.

For each dimension $d$ and value $v$, we compute the marginal lift:
\[
\text{lift}(d, v) = \mathbb{E}[\text{badness} \mid d = v] - \mathbb{E}[\text{badness}]
\]
where $\text{badness} = 1 - \text{mean\_llm\_judge\_score}$ and expectations are weighted by sample count.

For an untested (zero-support) full slice $s = (v_1, \ldots, v_K)$, we predict:
\[
\widehat{\text{badness}}(s) = \text{clip}\left(\mathbb{E}[\text{badness}] + \sum_{k=1}^{K} \text{lift}(d_k, v_k),\; 0,\; 1\right)
\]

Slices are ranked by predicted badness (descending).

\begin{figure}[t]
\centering
\caption{Marginal-Lift Model (SliceFinder-Inspired)}
\label{alg:marginal_lift}
\begin{minipage}{0.97\linewidth}
\small
\begin{lstlisting}[
  mathescape=true,
  basicstyle=\scriptsize\ttfamily,
  columns=fullflexible,
  frame=single,
  breaklines=true,
  aboveskip=0.5em,
  belowskip=0.5em
]
Require: Triage table $T$ with dimensions $\{d_1, \ldots, d_K\}$, weights $w_i$, scores $y_i$
Require: Untested full slices $\mathcal{U}$
Ensure: Ranked list of untested slices
1: $\bar{b} \gets \sum_i w_i (1 - y_i) / \sum_i w_i$  // global badness
2: for each dimension $d_k$ and value $v \in \text{vocab}(d_k)$ do
3:   $\bar{b}_{d_k=v} \gets \sum_{i: x_i[d_k]=v} w_i (1 - y_i) / \sum_{i: x_i[d_k]=v} w_i$
4:   $\text{lift}(d_k, v) \gets \bar{b}_{d_k=v} - \bar{b}$
5: for each untested slice $s = (v_1, \ldots, v_K) \in \mathcal{U}$ do
6:   $\widehat{b}(s) \gets \text{clip}\left(\bar{b} + \sum_{k=1}^{K} \text{lift}(d_k, v_k),\; 0,\; 1\right)$
7: return $\mathcal{U}$ sorted by $\widehat{b}(s)$ descending
\end{lstlisting}
\end{minipage}
\end{figure}

\subsection{Hyperparameters}

Table~\ref{tab:baseline_hyperparams} lists the hyperparameters used for each baseline.

\begin{table}[!ht]
\centering
\caption{Baseline adaptation hyperparameters.}
\label{tab:baseline_hyperparams}
\begin{tabular}{llrl}
\toprule
\textbf{Method} & \textbf{Parameter} & \textbf{Value} & \textbf{Description} \\
\midrule
SliceLine & $k$ & 1000--5000 & Number of predicates extracted \\
SliceLine & max\_l & 8 & Maximum predicate length (all dimensions) \\
SliceLine & min\_sup & 1 & Minimum support for predicates \\
SliceLine & $\alpha$ & 0.95 & Effect size threshold \\
\midrule
Marginal-Lift & rule\_length & 8 & Full slices only (all dimensions) \\
Marginal-Lift & top\_k & 1000--5000 & Number of ranked slices retained \\
\bottomrule
\end{tabular}
\end{table}

\subsection{Limitations of Adaptations}

We acknowledge that these adaptations are approximations:

\begin{itemize}[leftmargin=*]
    \item \textbf{SliceLine:} Partial predicates may over-generalize; a high-error predicate like \texttt{weather\_visibility=foggy} propagates its score to all foggy slices regardless of other dimensions. The specificity-first ranking is a heuristic without theoretical justification.
    \item \textbf{Marginal-Lift:} The additive decomposition ignores dimension interactions. For example, ``foggy + night'' may be substantially worse than the sum of marginal effects for foggy and night individually.
\end{itemize}

Despite these limitations, we believe these adaptations represent reasonable baselines for a task the original methods were not designed for. The fact that \textsc{SliceScorer} outperforms both adaptations suggests its design---combining coverage priors with neighbor-based failure propagation---is better suited to missing-slice ranking.

\section{LLM-as-a-Judge via VLMEvalKit}
\label{sec:llm-judge}

\paragraph{Description.} We use VLMEvalKit \cite{duan2024vlmevalkit} to score VLM predictions with an LLM judge (GPT-4.1 in our case) to decide whether each predicted answer is correct or incorrect. No image or video is sent to the judge; only text (question, options if MCQ, predicted answer, ground-truth answer) is passed.

\paragraph{Components.} We build the judge using VLMEvalKit's standard judge-construction helper and its packaged default system prompt, reproduced in \cref{lst:judge_system_prompt}. For the user prompt, we follow \cref{lst:judge_user_prompt_MCQ} for multiple-choice questions and \cref{lst:judge_user_prompt_VQA} for open-ended visual QA. For MCQ, the prompt includes the question text, the option list (formatted as ``A.~\ldots'', ``B.~\ldots''), the model's predicted answer, and the ground-truth answer. For open-ended VQA, it includes the question, the model's free-text response, and the reference answer text (including alias keys when provided by the benchmark).

\begin{lstlisting}[label=lst:judge_system_prompt,caption=LLM Judge System Prompt]
{"task": "Evaluate whether the answer to a question is correct.",
 "requirements": [
  "Compare an answer to a question with the ground truth answer. Determine whether it is correct.",
  "You must ignore any analysis of the problem if present. You must focus only on the final answer.",
  "You must answer in the following json format: {"verdict": "(1 for correct, 0 for incorrect)"}"
 ]}
\end{lstlisting}

\begin{lstlisting}[label=lst:judge_user_prompt_MCQ,caption=LLM Judge User Prompt (MCQ setting)]
You are an AI assistant evaluating a multiple-choice question answer.

Question: <question>

Options:
A. <option_0>
B. <option_1>
...

Predicted Answer: <predicted_answer>
Ground Truth Answer: <ground_truth_answer>

Please judge whether the predicted answer is correct. Respond with only "CORRECT" or "INCORRECT".
\end{lstlisting}

\begin{lstlisting}[label=lst:judge_user_prompt_VQA,caption=LLM Judge User Prompt (VQA or open-ended setting)]
You are an AI assistant evaluating a visual question answering response.

Question: <question>

Predicted Answer: <predicted_text>
Ground Truth Answer: <ground_truth_text>

Please judge whether the predicted answer is correct. Consider semantic equivalence - answers that convey the same meaning should be considered correct even if worded differently. Respond with only "CORRECT" or "INCORRECT".
\end{lstlisting}

\section{Dataset collection for evaluations in this paper}
\label{sec:dataset_collection}
We collected 15 datasets spanning diverse conditions and attributes to populate the oracle ranking table with broad slice coverage under the base and extended ODD definitions. \Cref{tab:sampled_set} summarizes the statistics of the seed evaluation table. The seed set was constructed with two key design choices: (1) each slice contains at least 10 examples to ensure statistical reliability, and (2) all QA annotations were human-generated to maximize label quality and support trustworthy downstream decisions. \Cref{tab:oracle_set} reports the statistics of the oracle ranking dataset pool used to evaluate all VLMs and construct the ground-truth failure triage. Importantly, there is no overlap in examples between the seed evaluation table and the ground-truth slice-based evaluation set.

\paragraph{Human verification of predicted labels.}
Four experts independently judged whether each model-proposed scene attribute and QA field was correct (\textit{Yes}/\textit{No}/\textit{Unsure}) on 100 stratified examples (same pool for all raters).
Pooled inter-rater agreement over all judgments yields Fleiss' $\kappa{=}0.2395$ and mean observed pairwise agreement $\bar P{=}0.8263$. Our dataset collection consists of both images and videos and covers multiple QA task types, including multiple-choice (MCQ) and open-ended VQA. To annotate pixel-level attributes aligned with the ODD dimensions, we use the Qwen3-30B model. For constructing QA pairs corresponding to the ODD \texttt{question\_category} axis, we proceed in two stages. First, we classify ground-truth QA pairs from datasets that already contain human-generated annotations. These labeled examples are then used as few-shot demonstrations to prompt GPT-5.2, which generates QA pairs for datasets that lack original QA annotations.  

\paragraph{Slice signature construction.}
Each test case (one visual input and its associated QA) is assigned a \emph{slice signature}: a single tuple over the same eight discrete ODD fields used in the main text (seven scene/context dimensions plus \texttt{question\_category}). The seven scene-related fields are produced by the Qwen3-30B vision--language annotator under \emph{forced choice}: for each field, the model must select exactly one label from that dimension's finite in-ODD vocabulary (there are no continuous raw measurements in the signature; discretization is implicit in this closed-set prediction). The \texttt{question\_category} label comes from dataset metadata when the QA is native to the benchmark; for QA pairs produced by the GPT-5.2 generation stage above, the category is assigned in that pipeline and mapped to the same closed \texttt{question\_category} set. For \emph{video} clips, the VLM is given the full temporal input and returns one global label per scene field (a single seven-tuple for the clip, not a separate tuple per frame). Test cases that share the same eight-tuple are grouped into the same slice for aggregation and scoring.

\begin{table}[!th]
\centering
\caption{\textbf{Seed Evaluation set.} Dataset, number of examples, QA type (MCQ/VQA), and number of unique slices.}
\label{tab:sampled_set}
\begin{tabular}{lllr}
\toprule
Dataset & \# Examples & QA type & \# Unique slices \\
\midrule
Context-VLM \cite{Shounak2024-rc} & 941 & MCQ & 156 \\
DRAMA \cite{malla2023drama} & 875 & MCQ & 131 \\
EPFL \cite{corbiere2025drivingvqa} & 70 & MCQ & 37 \\
LingoQA \cite{marcu2024lingoqa} & 19 & VQA & 13 \\
RealWorldVQA \cite{xai_realworldqa_2024} & 22 & VQA & 17 \\
SUTD \cite{xu2021sutd} & 451 & MCQ & 132 \\
TUMTraffic-VQA \cite{Zhou2025-fe} & 572 & MCQ & 118 \\
\bottomrule
\end{tabular}
\end{table}

\begin{table}[!ht]
\centering
\caption{\textbf{Oracle Evaluation Dataset.} Sample counts for datasets used in oracle.}
\label{tab:oracle_set}
\begin{tabular}{lr}
\toprule
Dataset & Count \\
\midrule
DRAMA \cite{malla2023drama} & 104{,}481 \\
Context-VLM \cite{Shounak2024-rc} & 34{,}267 \\
TUMTraffic-VQA \cite{Zhou2025-fe} & 24{,}309 \\
SUTD \cite{xu2021sutd} & 5{,}624 \\
LingoQA \cite{marcu2024lingoqa} & 962 \\
EPFL \cite{corbiere2025drivingvqa} & 929 \\
RealWorldVQA \cite{xai_realworldqa_2024} & 522 \\
Canadian Adverse Driving Conditions dataset \cite{pitropov2021canadian} & 6{,}899 \\
DAWN \cite{kenk2020dawn} & 1{,}027 \\
PESOTIF \cite{peng2023pesotif} & 1{,}036 \\
RADIATE \cite{sheeny2021radiate} & 32{,}997 \\
ZOD \cite{alibeigi2023zenseact} & 99{,}998 \\
WEDGE \cite{marathe2023wedge} & 3{,}360  \\
WildDash2 \cite{zendel2018wilddash} & 4{,}256 \\
Roadworks \cite{ghosh2025roadwork} & 8{,}549 \\ 
\bottomrule
\end{tabular}
\end{table}

\section{Datastore for retrieval with \textsc{SliceNav}}
\label{sec:datastore_retrieval_slicenav}

\textsc{SliceNav} uses a retrieval datastore to ground slice recommendations in existing driving footage before additional simulation or evaluation is requested. The datastore aggregates large-scale autonomous-driving video sources with complementary coverage: nuPlan ~\cite{Caesar2021nuPlan} contributes geographically diverse closed-loop driving logs , NVIDIA Alpamayo/Physical AI open datasets ~\cite{NVIDIAAlpamayo2025} contribute large-scale long-tail driving data for autonomous-vehicle development, and BDD100K \cite{Yu2020BDD100K} contributes diverse road-scene videos spanning weather, time, and scene conditions. Table~\ref{tab:slicenav_retrieval_datastore} summarizes the video inventory used for retrieval. If datastore retrieval returns no matching examples for a recommended slice, we generate backup synthetic examples for that slice using Qwen-Image~\cite{wu2025qwenimagetechnicalreport}.

\begin{table}[!ht]
\centering
\caption{\textbf{Retrieval datastore used by \textsc{SliceNav}.} Video counts are approximate and describe the source inventory available for retrieval.}
\label{tab:slicenav_retrieval_datastore}
\begin{tabular}{lr}
\toprule
\textbf{Dataset} & \textbf{Videos} \\
\midrule
nuPlan~\cite{Caesar2021nuPlan} & $\sim$14{,}000 \\
Alpamayo~\cite{NVIDIAAlpamayo2025} & 1{,}000{,}000 \\
BDD100K~\cite{Yu2020BDD100K} & 56{,}413 \\
\bottomrule
\end{tabular}
\end{table}

\section{Experimental protocols for Analysis 1--4}
\label{sec:experimental_protocols}

\textbf{Analysis~1 (in-ODD missing-slice ranking):} We rank missing in-ODD slices from the seed triage table (no user query) using \textsc{SliceScorer} and adapted baselines \cite{sagadeeva2021sliceline, chung2018slicefinder}, and evaluate with paired Regret@$K$ and ILD@$K$ against the oracle relevance table.

\textbf{Analysis~2 (outside-ODD ranking):} We rank slices from a shared expanded outside-ODD candidate pool with \textsc{SliceScorer}, adapted baselines, embedding-kNN, and random, and evaluate again with paired Regret@$K$ and ILD@$K$.

\textbf{Analysis~3 (qualitative workflow examples):} We run 15 developer queries through the full agent and report representative completed runs showing query interpretation, optional ODD expansion, deterministic slice ranking, and acquisition/evaluation chaining.

\textbf{Analysis~4 (reliability and stochasticity):} We repeat the 15-query set across 10 runs for both baseline LLM agent and \textsc{Slicenav}, and quantify run-to-run variation in returned slices (set-overlap Jaccard) (Appendix \ref{sec:analysis4_queries}).

\section{Analysis 4 query set (15 prompts)}
\label{sec:analysis4_queries}

For the Analysis 4 reliability/stochasticity study, we use the following 15 developer prompts and repeat each prompt across 10 runs.

\begin{enumerate}\item What driving scenarios outside our current ODD should we prioritize for testing?
\item Identify high-risk conditions we haven't validated yet.
\item What blind spots exist in our current test coverage?
\item Suggest untested edge cases that could cause failures in deployment.
\item We're expanding to construction zones - what untested scenarios should we prioritize?
\item What snow and ice scenarios are we missing?
\item What nighttime edge cases should we test?
\item We're deploying on rural highways - what scenarios haven't we covered?
\item What tunnel and underpass scenarios need testing?
\item We're concerned about heavy rain - what wet road scenarios should we prioritize?
\item What rush hour congestion scenarios are undertested?
\item What dawn and dusk visibility scenarios are we missing?
\item Find high-risk untested scenarios and acquire test samples for evaluation.
\item Identify missing foggy highway slices, retrieve test images, and evaluate the model.
\item Find untested night driving scenarios, generate synthetic test data, and run VLM evaluation.
\end{enumerate}

\section{LLM choose-slices baseline prompts}
\label{sec:llm_choose_slices_prompts}

This section reproduces the prompts used by the LLM baseline in Analysis 4. Note that the baseline conditions on a \emph{compact text summary} of the seed triage CSV - marginals, support statistics, and aggregate judge scores - not the verbatim full per-row table in the prompt. At our seed-inventory scales, embedding the complete table would exceed practical context-length budgets for general-purpose chat models, so verbatim full-table conditioning is infeasible for this listing baseline. Variance and request-vs.-fulfillment analyses for LLM listing should be read under this summary-conditioned regime. Repeated listing calls use the API \emph{default} decoding settings (we do not set temperature or sampling knobs to deterministic values), so run-to-run variability reflects ordinary chat-model behavior rather than a controlled greedy decode.

\begin{lstlisting}[label=lst:choose_slices_system_prompt,caption=Choose-slices baseline system prompt]
You are an expert analyzing VLM (Vision-Language Model) failure patterns in autonomous driving scenarios.

Your task is to choose which outside-ODD slices you recommend for testing.

"Outside-ODD" means: at least one attribute value in the slice is in the extended ODD but NOT in the base ODD (i.e. not yet covered by triage).

You will be given:
- A summary of triage data (tested slices and performance)
- Base ODD and extended ODD definitions

Respond with ONLY a CSV of your chosen outside-ODD slices. Do not generate any code. Use the exact dimension names as column headers. One row per slice.
\end{lstlisting}

\begin{lstlisting}[label=lst:choose_slices_instructions,caption=Choose-slices baseline instructions prompt]
## Triage Summary (observed slices inside ODD)

{triage_summary}

## Slice Dimensions (use these exact names as CSV columns)

{slice_cols}

## Base ODD Values (inside ODD - already tested)

{base_odd}

## Extended ODD Values (superset - choose outside-ODD slices from these)

{extended_odd}

## Task

Choose the outside-ODD slices you recommend for testing. Each chosen slice must have at least one attribute value that is in the extended ODD but NOT in the base ODD.

Output your chosen slices as CSV only:
- First line: header with exactly these columns (comma-separated): {csv_header}
- Following lines: one row per slice, same column order, comma-separated values
- Strictly Output {top_k} slices.

Do not include any explanation or code. Only the CSV (you may wrap it in a ```csv code block).
\end{lstlisting}

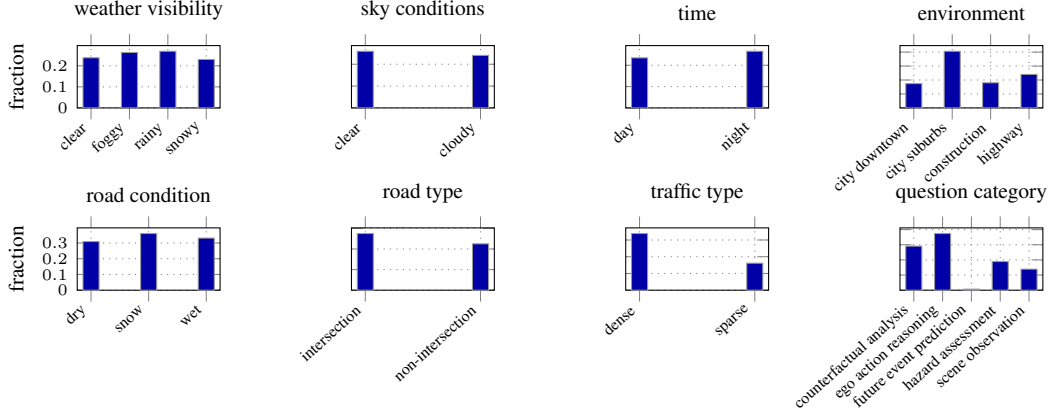
\begin{figure}[t]
\centering
\resizebox{\linewidth}{!}{%
\begin{tikzpicture}
\begin{groupplot}[
  group style={
    group size=4 by 2,
    horizontal sep=1.725cm,
    vertical sep=1.575cm,
    y descriptions at=edge left,
  },
  width=0.2475\textwidth,
  height=2.4cm,
  enlarge x limits=0.12,
  ymin=0,
  every axis/.append style={ybar, bar width=6pt},
]
\nextgroupplot[
  title={\footnotesize weather visibility},
  ymin=0,
  xtick={1,2,3,4},
  xticklabels={{clear},{foggy},{rainy},{snowy}},
  x tick label style={rotate=42,anchor=east,font=\tiny},
  yticklabel style={font=\tiny},
  title style={font=\small},
  grid=major,
  major grid style={dotted,gray,thin},ylabel={\footnotesize fraction}
] \addplot+[ybar,fill=blue!65!black,draw=black!40] coordinates {(1,0.238000) (2,0.263000) (3,0.269000) (4,0.230000)};

\nextgroupplot[
  title={\footnotesize sky conditions},
  ymin=0,
  xtick={1,2},
  xticklabels={{clear},{cloudy}},
  x tick label style={rotate=42,anchor=east,font=\tiny},
  yticklabel style={font=\tiny},
  title style={font=\small},
  grid=major,
  major grid style={dotted,gray,thin}
] \addplot+[ybar,fill=blue!65!black,draw=black!40] coordinates {(1,0.519000) (2,0.481000)};

\nextgroupplot[
  title={\footnotesize time},
  ymin=0,
  xtick={1,2},
  xticklabels={{day},{night}},
  x tick label style={rotate=42,anchor=east,font=\tiny},
  yticklabel style={font=\tiny},
  title style={font=\small},
  grid=major,
  major grid style={dotted,gray,thin}
] \addplot+[ybar,fill=blue!65!black,draw=black!40] coordinates {(1,0.469000) (2,0.531000)};

\nextgroupplot[
  title={\footnotesize environment},
  ymin=0,
  xtick={1,2,3,4},
  xticklabels={{city downtown},{city suburbs},{construction},{highway}},
  x tick label style={rotate=42,anchor=east,font=\tiny},
  yticklabel style={font=\tiny},
  title style={font=\small},
  grid=major,
  major grid style={dotted,gray,thin}
] \addplot+[ybar,fill=blue!65!black,draw=black!40] coordinates {(1,0.174000) (2,0.406000) (3,0.180000) (4,0.240000)};

\nextgroupplot[
  title={\footnotesize road condition},
  ymin=0,
  xtick={1,2,3},
  xticklabels={{dry},{snow},{wet}},
  x tick label style={rotate=42,anchor=east,font=\tiny},
  yticklabel style={font=\tiny},
  title style={font=\small},
  grid=major,
  major grid style={dotted,gray,thin},ylabel={\footnotesize fraction}
] \addplot+[ybar,fill=blue!65!black,draw=black!40] coordinates {(1,0.309000) (2,0.360000) (3,0.331000)};

\nextgroupplot[
  title={\footnotesize road type},
  ymin=0,
  xtick={1,2},
  xticklabels={{intersection},{non-intersection}},
  x tick label style={rotate=42,anchor=east,font=\tiny},
  yticklabel style={font=\tiny},
  title style={font=\small},
  grid=major,
  major grid style={dotted,gray,thin}
] \addplot+[ybar,fill=blue!65!black,draw=black!40] coordinates {(1,0.550000) (2,0.450000)};

\nextgroupplot[
  title={\footnotesize traffic type},
  ymin=0,
  xtick={1,2},
  xticklabels={{dense},{sparse}},
  x tick label style={rotate=42,anchor=east,font=\tiny},
  yticklabel style={font=\tiny},
  title style={font=\small},
  grid=major,
  major grid style={dotted,gray,thin}
] \addplot+[ybar,fill=blue!65!black,draw=black!40] coordinates {(1,0.677000) (2,0.323000)};

\nextgroupplot[
  title={\footnotesize question category},
  ymin=0,
  xtick={1,2,3,4,5},
  xticklabels={{counterfactual analysis},{ego action reasoning},{future event prediction},{hazard assessment},{scene observation}},
  x tick label style={rotate=42,anchor=east,font=\tiny},
  yticklabel style={font=\tiny},
  title style={font=\small},
  grid=major,
  major grid style={dotted,gray,thin}
] \addplot+[ybar,fill=blue!65!black,draw=black!40] coordinates {(1,0.291000) (2,0.374000) (3,0.007000) (4,0.189000) (5,0.139000)};

\end{groupplot}
\end{tikzpicture}%
}
\caption{\textbf{WiseAD SliceScorer recommendation profile.} Distribution of top-ranked missing-slice recommendations across ODD attributes.}
\label{fig:slicescorer_rec_profile_wisead}
\end{figure}

\begin{figure}[t]
\centering
\resizebox{\linewidth}{!}{%
\begin{tikzpicture}
\begin{groupplot}[
  group style={
    group size=4 by 2,
    horizontal sep=1.725cm,
    vertical sep=1.575cm,
    y descriptions at=edge left,
  },
  width=0.2475\textwidth,
  height=2.4cm,
  enlarge x limits=0.12,
  ymin=0,
  every axis/.append style={ybar, bar width=6pt},
]
\nextgroupplot[
  title={\footnotesize weather visibility},
  ymin=0,
  xtick={1,2,3,4},
  xticklabels={{clear},{foggy},{rainy},{snowy}},
  x tick label style={rotate=42,anchor=east,font=\tiny},
  yticklabel style={font=\tiny},
  title style={font=\small},
  grid=major,
  major grid style={dotted,gray,thin},ylabel={\footnotesize fraction}
] \addplot+[ybar,fill=orange!85!black,draw=black!40] coordinates {(1,0.224000) (2,0.259000) (3,0.317000) (4,0.200000)};

\nextgroupplot[
  title={\footnotesize sky conditions},
  ymin=0,
  xtick={1,2},
  xticklabels={{clear},{cloudy}},
  x tick label style={rotate=42,anchor=east,font=\tiny},
  yticklabel style={font=\tiny},
  title style={font=\small},
  grid=major,
  major grid style={dotted,gray,thin}
] \addplot+[ybar,fill=orange!85!black,draw=black!40] coordinates {(1,0.495000) (2,0.505000)};

\nextgroupplot[
  title={\footnotesize time},
  ymin=0,
  xtick={1,2},
  xticklabels={{day},{night}},
  x tick label style={rotate=42,anchor=east,font=\tiny},
  yticklabel style={font=\tiny},
  title style={font=\small},
  grid=major,
  major grid style={dotted,gray,thin}
] \addplot+[ybar,fill=orange!85!black,draw=black!40] coordinates {(1,0.492000) (2,0.508000)};

\nextgroupplot[
  title={\footnotesize environment},
  ymin=0,
  xtick={1,2,3,4},
  xticklabels={{city downtown},{city suburbs},{construction},{highway}},
  x tick label style={rotate=42,anchor=east,font=\tiny},
  yticklabel style={font=\tiny},
  title style={font=\small},
  grid=major,
  major grid style={dotted,gray,thin}
] \addplot+[ybar,fill=orange!85!black,draw=black!40] coordinates {(1,0.327000) (2,0.268000) (3,0.205000) (4,0.200000)};

\nextgroupplot[
  title={\footnotesize road condition},
  ymin=0,
  xtick={1,2,3},
  xticklabels={{dry},{snow},{wet}},
  x tick label style={rotate=42,anchor=east,font=\tiny},
  yticklabel style={font=\tiny},
  title style={font=\small},
  grid=major,
  major grid style={dotted,gray,thin},ylabel={\footnotesize fraction}
] \addplot+[ybar,fill=orange!85!black,draw=black!40] coordinates {(1,0.326000) (2,0.321000) (3,0.353000)};

\nextgroupplot[
  title={\footnotesize road type},
  ymin=0,
  xtick={1,2},
  xticklabels={{intersection},{non-intersection}},
  x tick label style={rotate=42,anchor=east,font=\tiny},
  yticklabel style={font=\tiny},
  title style={font=\small},
  grid=major,
  major grid style={dotted,gray,thin}
] \addplot+[ybar,fill=orange!85!black,draw=black!40] coordinates {(1,0.522000) (2,0.478000)};

\nextgroupplot[
  title={\footnotesize traffic type},
  ymin=0,
  xtick={1,2},
  xticklabels={{dense},{sparse}},
  x tick label style={rotate=42,anchor=east,font=\tiny},
  yticklabel style={font=\tiny},
  title style={font=\small},
  grid=major,
  major grid style={dotted,gray,thin}
] \addplot+[ybar,fill=orange!85!black,draw=black!40] coordinates {(1,0.489000) (2,0.511000)};

\nextgroupplot[
  title={\footnotesize question category},
  ymin=0,
  xtick={1,2,3,4,5},
  xticklabels={{counterfactual analysis},{ego action reasoning},{future event prediction},{hazard assessment},{scene observation}},
  x tick label style={rotate=42,anchor=east,font=\tiny},
  yticklabel style={font=\tiny},
  title style={font=\small},
  grid=major,
  major grid style={dotted,gray,thin}
] \addplot+[ybar,fill=orange!85!black,draw=black!40] coordinates {(1,0.709000) (2,0.000000) (3,0.062000) (4,0.227000) (5,0.002000)};

\end{groupplot}
\end{tikzpicture}%
}
\caption{\textbf{DriveMM SliceScorer recommendation profile.} Distribution of top-ranked missing-slice recommendations across ODD attributes.}
\label{fig:slicescorer_rec_profile_drivemm}
\end{figure}

\begin{figure}[t]
\centering
\resizebox{\linewidth}{!}{%
\begin{tikzpicture}
\begin{groupplot}[
  group style={
    group size=4 by 2,
    horizontal sep=1.725cm,
    vertical sep=1.575cm,
    y descriptions at=edge left,
  },
  width=0.2475\textwidth,
  height=2.4cm,
  enlarge x limits=0.12,
  ymin=0,
  every axis/.append style={ybar, bar width=6pt},
]
\nextgroupplot[
  title={\footnotesize weather visibility},
  ymin=0,
  xtick={1,2,3,4},
  xticklabels={{clear},{foggy},{rainy},{snowy}},
  x tick label style={rotate=42,anchor=east,font=\tiny},
  yticklabel style={font=\tiny},
  title style={font=\small},
  grid=major,
  major grid style={dotted,gray,thin},ylabel={\footnotesize fraction}
] \addplot+[ybar,fill=green!55!black,draw=black!40] coordinates {(1,0.262000) (2,0.263000) (3,0.271000) (4,0.204000)};

\nextgroupplot[
  title={\footnotesize sky conditions},
  ymin=0,
  xtick={1,2},
  xticklabels={{clear},{cloudy}},
  x tick label style={rotate=42,anchor=east,font=\tiny},
  yticklabel style={font=\tiny},
  title style={font=\small},
  grid=major,
  major grid style={dotted,gray,thin}
] \addplot+[ybar,fill=green!55!black,draw=black!40] coordinates {(1,0.516000) (2,0.484000)};

\nextgroupplot[
  title={\footnotesize time},
  ymin=0,
  xtick={1,2},
  xticklabels={{day},{night}},
  x tick label style={rotate=42,anchor=east,font=\tiny},
  yticklabel style={font=\tiny},
  title style={font=\small},
  grid=major,
  major grid style={dotted,gray,thin}
] \addplot+[ybar,fill=green!55!black,draw=black!40] coordinates {(1,0.479000) (2,0.521000)};

\nextgroupplot[
  title={\footnotesize environment},
  ymin=0,
  xtick={1,2,3,4},
  xticklabels={{city downtown},{city suburbs},{construction},{highway}},
  x tick label style={rotate=42,anchor=east,font=\tiny},
  yticklabel style={font=\tiny},
  title style={font=\small},
  grid=major,
  major grid style={dotted,gray,thin}
] \addplot+[ybar,fill=green!55!black,draw=black!40] coordinates {(1,0.311000) (2,0.275000) (3,0.183000) (4,0.231000)};

\nextgroupplot[
  title={\footnotesize road condition},
  ymin=0,
  xtick={1,2,3},
  xticklabels={{dry},{snow},{wet}},
  x tick label style={rotate=42,anchor=east,font=\tiny},
  yticklabel style={font=\tiny},
  title style={font=\small},
  grid=major,
  major grid style={dotted,gray,thin},ylabel={\footnotesize fraction}
] \addplot+[ybar,fill=green!55!black,draw=black!40] coordinates {(1,0.313000) (2,0.340000) (3,0.347000)};

\nextgroupplot[
  title={\footnotesize road type},
  ymin=0,
  xtick={1,2},
  xticklabels={{intersection},{non-intersection}},
  x tick label style={rotate=42,anchor=east,font=\tiny},
  yticklabel style={font=\tiny},
  title style={font=\small},
  grid=major,
  major grid style={dotted,gray,thin}
] \addplot+[ybar,fill=green!55!black,draw=black!40] coordinates {(1,0.526000) (2,0.474000)};

\nextgroupplot[
  title={\footnotesize traffic type},
  ymin=0,
  xtick={1,2},
  xticklabels={{dense},{sparse}},
  x tick label style={rotate=42,anchor=east,font=\tiny},
  yticklabel style={font=\tiny},
  title style={font=\small},
  grid=major,
  major grid style={dotted,gray,thin}
] \addplot+[ybar,fill=green!55!black,draw=black!40] coordinates {(1,0.501000) (2,0.499000)};

\nextgroupplot[
  title={\footnotesize question category},
  ymin=0,
  xtick={1,2,3,4,5},
  xticklabels={{counterfactual analysis},{ego action reasoning},{future event prediction},{hazard assessment},{scene observation}},
  x tick label style={rotate=42,anchor=east,font=\tiny},
  yticklabel style={font=\tiny},
  title style={font=\small},
  grid=major,
  major grid style={dotted,gray,thin}
] \addplot+[ybar,fill=green!55!black,draw=black!40] coordinates {(1,0.616000) (2,0.175000) (3,0.015000) (4,0.194000) (5,0.000000)};

\end{groupplot}
\end{tikzpicture}%
}
\caption{\textbf{Cosmos SliceScorer recommendation profile.} Distribution of top-ranked missing-slice recommendations across ODD attributes.}
\label{fig:slicescorer_rec_profile_cosmos}
\end{figure}
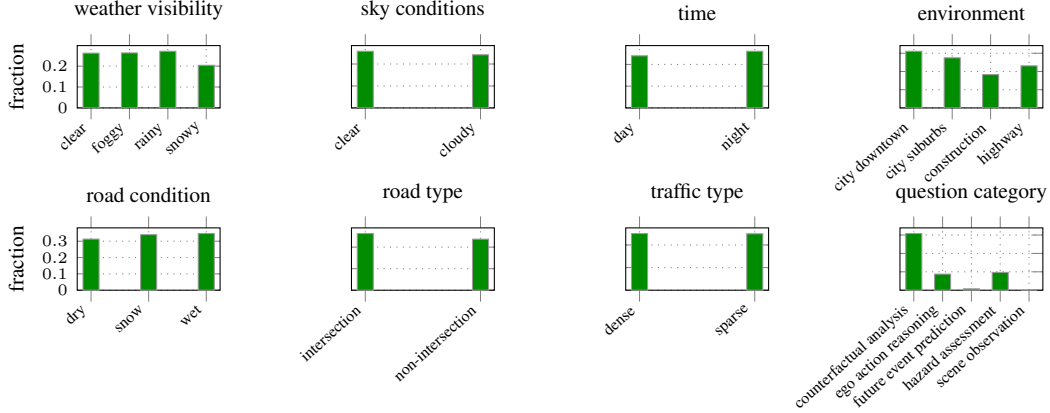

\section{Full rationales for qualitative workflow examples}
\label{sec:qualitative_rationales}

\paragraph{Broad outside-ODD query (Example A).}
For the broad prompt, \textsc{SliceNav} first inspected the current triage table and found that failures concentrate around rainy conditions, intersections, hazard assessment, ego action reasoning, and counterfactual reasoning. Because the user asked for scenarios outside the current ODD, the controller expanded the ontology before ranking missing slices. The selected additions expose distribution shifts adjacent to the observed failure modes: \texttt{dawn} and \texttt{dusk} test transitional lighting, \texttt{puddles}, \texttt{muddy}, and \texttt{icy} test road-surface states beyond \texttt{dry}/\texttt{wet}/\texttt{snow}, and \texttt{light\_rain}/\texttt{heavy\_rain} refine the coarse \texttt{rainy} bucket. The top recommendations therefore prioritize downtown intersections at dusk with dry or wet surfaces, followed by nighttime puddle/muddy variants, rather than listing arbitrary exotic conditions.

\paragraph{Focused outside-ODD query (Example B).}
For the tunnel/underpass prompt, the query itself supplied a target deployment environment. The controller therefore skipped broad LLM vocabulary search and inserted the explicit new value \texttt{tunnel} into the environment dimension. The outside-ODD missing-slice ranker then returned top slices all using \texttt{environment=tunnel}, concentrated on foggy daytime dense-traffic cases while varying road condition and road type. A separate failure-summary pass found no tunnel or underpass groups in the existing triage failures, supporting the interpretation that these are coverage gaps rather than already-tested weak spots. Underpass did not appear in the top recommendations, so the paper reports the evidenced tunnel result instead of inventing unsupported underpass slices.

\paragraph{Full acquisition loop (Example C).}
For the night-driving acquisition prompt, the controller stayed inside the base ODD because the requested value \texttt{night} already exists in the taxonomy. It ranked missing night slices, then chained the recommendation to synthetic data generation and VLM evaluation. The highest-priority gaps were rainy nighttime city-suburb intersections for hazard assessment, varying road surface and traffic density, with additional night gaps for counterfactual highway reasoning and dense downtown ego-action reasoning. The run generated 10 synthetic samples, one per candidate slice, and evaluated Cosmos-Reason2-2B. Restricting the summary to the 8 night slices, all categories had MCQ accuracy 0.0; the worst slice was rainy/cloudy/night/city-suburbs/wet/intersection/sparse/hazard-assessment, with MCQ accuracy 0.0, VQA accuracy 0.0, and mean judge score 0.0.

\subsection{Additional qualitative examples}
\label{sec:additional_qualitative_examples}

\paragraph{Generic high-risk acquisition (Example D).}
\textbf{Query:} ``Find high-risk untested scenarios and acquire test samples for evaluation.'' \textbf{Tools:} within-ODD missing-slice ranking $\rightarrow$ datastore retrieval $\rightarrow$ simulation fallback for uncovered slices $\rightarrow$ dataset inspection. \textbf{Vocabulary:} no extension; this run stayed within the base ODD. \textbf{Top-2 slices:} (rainy, cloudy, night, city suburbs, wet, intersection, sparse, hazard assessment) and (rainy, cloudy, night, city suburbs, snow, intersection, sparse, hazard assessment). \textbf{Rationale:} the run shows that acquisition is a workflow decision, not only a ranking output: most slices were acquired from the datastore, while a missing retrieval was filled by simulation. The response also flags slice-fidelity risks from returned data fields, motivating post-filtering before treating retrieved samples as valid slice coverage.

\paragraph{Foggy-highway retrieval and evaluation (Example E).}
\textbf{Query:} ``Identify missing foggy highway slices, retrieve test images, and evaluate the model.'' \textbf{Tools:} foggy-highway gap enumeration $\rightarrow$ within-ODD slice ranking $\rightarrow$ datastore retrieval $\rightarrow$ VLM evaluation $\rightarrow$ condition-level analysis. \textbf{Vocabulary:} no extension; the query constrains existing ODD values. \textbf{Top-2 intended slices:} (foggy, cloudy, day, highway, dry, intersection, sparse, ego action reasoning) and (foggy, cloudy, day, highway, dry, intersection, sparse, hazard assessment). \textbf{Rationale:} the combinatorial audit finds 225 missing foggy-highway combinations out of 240 possible combinations. The retrieval and evaluation step is especially useful as evidence because the returned samples collapse to only two realized non-intersection dry daytime foggy-highway slices with missing question category, showing why acquisition workflows need slice-fidelity checks in addition to ranking.

\paragraph{Snow and ice coverage (Example F).}
\textbf{Query:} ``What snow and ice scenarios are we missing?'' \textbf{Tools:} within-ODD missing-slice ranking $\rightarrow$ refined top-$K$ snow query $\rightarrow$ existing-failure summary. \textbf{Vocabulary:} no automatic extension; the artifact notes that \texttt{ice} is not an explicit road-condition value, so \texttt{road\_condition=snow} is used as the closest snow/ice proxy. \textbf{Top-2 slices:} (rainy, cloudy, night, city suburbs, snow, intersection, sparse, hazard assessment) and (clear, cloudy, night, city downtown, snow, intersection, dense, ego action reasoning). \textbf{Rationale:} the run surfaces mixed-precipitation and snowy-surface gaps at intersections, then separates hazard assessment gaps from a secondary ego-action reasoning gap in dense urban night traffic.

\paragraph{Dawn/dusk illumination extension (Example G).}
\textbf{Query:} ``What dawn and dusk visibility scenarios are we missing?'' \textbf{Tools:} outside-ODD vocabulary expansion $\rightarrow$ outside-ODD missing-slice ranking $\rightarrow$ targeted inspection of dawn/dusk outputs. \textbf{Vocabulary:} \texttt{+dawn} and \texttt{+dusk} are added to the \texttt{time} dimension. \textbf{Top-2 slices:} (clear, clear, dusk, city downtown, dry, intersection, sparse) and (clear, clear, dusk, city downtown, wet, intersection, sparse). \textbf{Rationale:} unlike the broad outside-ODD query, this focused prompt produces a narrow twilight cluster: clear downtown sparse-traffic scenes at dawn or dusk, varying road surface and road geometry. The returned table omits question category, so the rationale explicitly treats this as a scene-condition coverage gap.

\paragraph{Construction-zone expansion (Example H).}
\textbf{Query:} ``We're expanding to construction zones -- what untested scenarios should we prioritize?'' \textbf{Tools:} within-ODD missing-slice ranking $\rightarrow$ existing-failure summary $\rightarrow$ construction-specific gap analysis. \textbf{Vocabulary:} no extension; \texttt{environment=construction} already exists in the base ODD. \textbf{Top-2 slices:} (rainy, cloudy, night, construction, snow, intersection, dense, hazard assessment) and (rainy, clear, night, construction, wet, intersection, dense, hazard assessment). \textbf{Rationale:} the construction-specific audit finds a large gap: 26 evaluated construction slices versus 934 untested construction combinations. The prioritized slices combine adverse weather, night driving, dense traffic, intersections, and hazard assessment, matching the safety-critical conditions a deployment expansion should test first.

\section{Slice-to-executable-test validation pipeline}
\label{sec:slice_to_executable_flow}

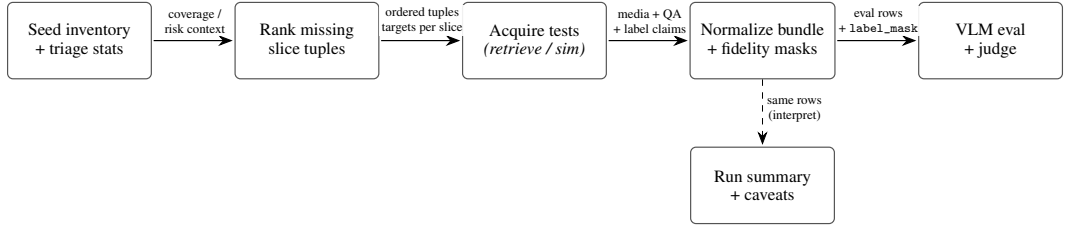
\begin{figure}[t]
\centering
\resizebox{\linewidth}{!}{%
\begin{tikzpicture}[
  node distance=5mm and 7mm,
  box/.style={rectangle, draw=gray!65!black, line width=0.55pt, rounded corners=2.5pt,
    align=center, inner sep=5pt, minimum height=1.25cm, minimum width=2.35cm, font=\footnotesize},
  arrow/.style={-{Stealth[length=2.4mm]}, line width=0.55pt, shorten >=0.8pt, shorten <=0.8pt},
  dasharrow/.style={-{Stealth[length=2.2mm]}, line width=0.45pt, dashed, shorten >=0.8pt},
  el/.style={font=\tiny, align=center, inner sep=1pt, fill=white, fill opacity=0.92, text opacity=1}
]

\node[box] (inv) {Seed inventory\\[-1pt]+ triage stats};
\node[box, right=1.35cm of inv] (rank) {Rank missing\\[-1pt]slice tuples};
\node[box, right=1.35cm of rank] (acq) {Acquire tests\\[-1pt]\textit{(retrieve / sim)}};
\node[box, right=1.35cm of acq] (bun) {Normalize bundle\\[-1pt]+ fidelity masks};
\node[box, right=1.35cm of bun] (vlm) {VLM eval\\[-1pt]+ judge};

\node[box, below=1.1cm of bun] (sum) {Run summary\\[-1pt]+ caveats};

\draw[arrow] (inv) -- node[el, above=2pt, pos=0.52] {coverage /\\risk context} (rank);
\draw[arrow] (rank) -- node[el, above=2pt, pos=0.52] {ordered tuples\\targets per slice} (acq);
\draw[arrow] (acq) -- node[el, above=2pt, pos=0.52] {media + QA\\+ label claims} (bun);
\draw[arrow] (bun) -- node[el, above=2pt, pos=0.52] {eval rows\\+ \texttt{label\_mask}} (vlm);

\draw[dasharrow] (bun) -- node[el, right=1pt, pos=0.45] {same rows\\(interpret)} (sum);

\end{tikzpicture}%
}
\caption{\textbf{Slice-to-executable-test flow.} Arrows show what each stage \emph{hands off} to the next; the dashed branch is narrative interpretation of the normalized bundle (qualitative Example~D: generic high-risk acquisition; Appendix \ref{sec:qualitative_rationales}).}
\label{fig:slice_to_exec_flow}
\end{figure}

\Cref{fig:slice_to_exec_flow} shows how a ranked slice recommendation is converted into an executable evaluation bundle. \textsc{SliceScorer} first ranks missing target slices from the seed inventory and triage statistics (estimated from seed evaluation set). For each target slice, the acquisition stage retrieves candidate media and QA pairs from the datastore in Appendix~\ref{sec:datastore_retrieval_slicenav}, with simulation used as a fallback when retrieval is insufficient. Retrieved and simulated candidates are then normalized into a common evaluation schema and annotated with slice-fidelity indicators that compare their metadata against the requested slice attributes. Candidates with missing, ambiguous, or mismatched metadata are retained for auditability but flagged separately, so they are not silently counted as valid evidence for the target slice. The final stage runs the VLM and judge on the normalized bundle and reports both evaluation results and acquisition caveats, including metadata gaps, road-type mismatches, or question-category mismatches.

\paragraph{Exemplar artifacts per stage.} An important purpose of our \textsc{SliceNav} pipeline is to make the transition from slice recommendation to executable testing explicit and auditable. In Example~D, the seed inventory and triage statistics define the current coverage and risk context. The ranked-slice stage produces \texttt{suggest\_top\_missing\_slices\_within\_ODD.csv}, whose leading target is a rainy, cloudy, night, city-suburbs, wet-road, intersection, sparse-traffic, hazard-assessment slice. The acquisition stage writes candidate media and QA items to \texttt{dataset\_retrieval.jsonl}; one retrieved QA item asks, ``What actions are currently taking place in the scene?'' The normalization stage formats these candidates for VLM evaluation and attaches fidelity annotations indicating whether each candidate matches the requested slice attributes. The VLM-evaluation stage runs the configured model and judge on the normalized bundle. Finally, \texttt{agent\_response.md} summarizes the recommended slices, evaluation outputs, and caveats about acquisition fidelity.

\section{Limitations and Broader Impacts}

\textbf{Limitations.} While \textsc{SliceNav} systematically identifies critical testing gaps, it depends on the granularity and comprehensiveness of the underlying ODD taxonomy and the seed evaluation table. If the taxonomy lacks certain environmental factors or if the seed table is excessively sparse, the \textsc{SliceScorer} might struggle to extrapolate neighbor failures reliably. Furthermore, while the LLM-driven candidate generation is constrained to logical ODD extensions, its output quality depends on the underlying LLM's reasoning capabilities.

\textbf{Broader Impacts.} By providing a structured, traceable framework to uncover "unknown unknown" failures in driving VLMs, our work supports systematic evaluation of scene understanding components increasingly used in autonomous driving stacks. Improved VLM evaluation methodologies help identify conditions where perception may degrade, informing downstream safety validation efforts. However, developers must not treat \textsc{SliceScorer} as a definitive guarantee of safety; it is a supplementary recommendation tool intended to prioritize testing efforts, not a complete proof of ODD coverage.

\stopcontents[appendix]
\FloatBarrier

\end{document}